\documentclass{article}

\usepackage{microtype}
\usepackage{graphicx}
\usepackage{multirow}
\usepackage{caption}
\usepackage{subcaption}
\usepackage{booktabs} 
\usepackage{enumitem}
\usepackage{mathtools}
\usepackage{amsmath,amssymb,stmaryrd}

\usepackage{pifont}
\newcommand{\cmark}{\ding{51}}%
\newcommand{\xmark}{\ding{55}}

\usepackage{hyperref}

\usepackage[accepted]{icml2020}

\icmltitlerunning{Semi-Supervised StyleGAN for Disentanglement Learning}

\begin{document}

\twocolumn[
\icmltitle{Semi-Supervised StyleGAN for Disentanglement Learning}




\begin{icmlauthorlist}
\icmlauthor{Weili Nie{$^*$}}{ri}
\icmlauthor{Tero Karras}{ke}
\icmlauthor{Animesh Garg}{ke,to}
\icmlauthor{Shoubhik Debnath}{ke}
\icmlauthor{Anjul Patney}{ke}

\icmlauthor{Ankit B. Patel}{ri}
\icmlauthor{Anima Anandkumar}{ke,ca}
\end{icmlauthorlist}

\icmlaffiliation{ri}{Rice University ({$^*$}Work done as a part of internship at Nvidia)}
\icmlaffiliation{ke}{Nvidia}
\icmlaffiliation{to}{University of Toronto}
\icmlaffiliation{ca}{California Institute of Technology}

\icmlcorrespondingauthor{Weili Nie}{wn8@rice.edu}


\icmlkeywords{Machine Learning, ICML}

\vskip 0.3in
]



\printAffiliationsAndNotice{}  

\begin{abstract}
Disentanglement learning is crucial for obtaining disentangled representations and controllable generation.
Current disentanglement methods face several inherent limitations: difficulty with high-resolution images, primarily focusing on learning disentangled representations, and non-identifiability due to the unsupervised setting.
To alleviate these limitations, we design new architectures and loss functions based on StyleGAN \citep{karras2019style}, for semi-supervised high-resolution disentanglement learning.
We create two complex high-resolution synthetic datasets for systematic testing.
We investigate the impact of limited supervision and find that using only 0.25\%$\sim$2.5\% of labeled data is sufficient for good disentanglement on both synthetic and real datasets. 
We propose new metrics to quantify generator controllability, and observe there may exist a crucial trade-off between disentangled representation learning and controllable generation.  
We also consider semantic fine-grained image editing to achieve better generalization to unseen images.


\end{abstract}

\vspace{-4mm}
\section{Introduction}
\label{submission}


Disentanglement learning with deep generative models has  attracted much attention recently~\citep{chen2016infogan, higgins2017beta, locatello2019challenging}. This is crucial for controllable generation, where the style codes specified to the generator need to separately control various factors of variation for faithful generation. Another goal is learning disentangled representations where the input samples can be encoded to latent factors that are disentangled. This has been argued as a key to  success of deep learning~\citep{achille2018emergence}. Previous works have primarily focused on only one of the above two objectives. Ideally, the ultimate goal of disentanglement learning is to achieve both the objectives at the same time, especially on more complex high-resolution images, and we pursue this goal in this paper. We first list the three main limitations of the current disentanglement methods.

First, much effort has   focused on unsupervised disentanglement methods \citep{chen2016infogan, higgins2017beta, nguyen2019hologan}. This is because a large number of fully annotated samples is expensive to obtain.
These methods suffer from  \textit{non-identifiability}, which means multiple repeated runs will not reliably observe the same latent representations~\citep{hyvarinen1999nonlinear,locatello2019challenging}. In addition, human feedback is needed   to discern (i) what factors of variation the model has learnt (e.g., object shape and color), and (ii) what semantic meaning different values of the discovered factor code represent (e.g., red and blue in a color factor code). 
To reliably control generation for practical use, adding a  small amount of labeled data may resolve the non-identifiability issue and lead to interpretable factors. Hence, we investigate the impact of \textit{limited supervision} on disentanglement learning.

Second, current disentanglement methods \citep{locatello2019challenging, locatello2019disentangling} are mainly developed and evaluated on relatively simple low-resolution images, such as dSprites \citep{dsprites17} and 3DShapes \citep{kim2018disentangling}, which raises concerns about their ability to scale up to more diverse, higher-resolution images.
For example, the use of 3D representations to disentangle the 3D pose may not easily apply to high-resolution images due to the computational cost. The difficulty of some deep generative models   at generating realistic images also limits their application in more complex  domains.
Furthermore, although there exist many real image datasets of high resolution, the latent factors are typically only partially observed or unbalanced, which makes it hard to scientifically study disentanglement. 
To gain practically useful insights, it is critical to first test disentanglement methods on complex, high-resolution, synthetic images wherein ground-truth factors are easy to obtain.

Third, most previous works \citep{kim2018disentangling,chen2018isolating} have primarily focused on learning disentangled representations by quantifying encoder disentanglement quality, in the hopes that a better disentangled encoder might also lead to a better disentangled generator.
However, to the best of our knowledge, there is no clear evidence supporting that a proportional relationship between encoder and generator disentanglement quality always exists. 
A good analogy is that an art critic can disentangle various painting styles and skills, but may not be able to create a good painting by combining these styles and skills.
Thus, results based solely on evaluating encoder disentanglement may be misleading, especially in tasks where the requirement for controllable generation is more critical.
This highlights the importance of measuring the generator disentanglement quality in order to properly evaluate and compare different methods.



\vspace{-2mm}
\paragraph{Main contributions. }
In this work, we investigate semi-supervised disentanglement learning based on StyleGAN \citep{karras2019style}, one of the state-of-the-art generative adversarial networks (GANs), for complex high-resolution images. In summary, our main contributions are as follows,
\begin{itemize} [topsep=0pt,itemsep=0ex,partopsep=1ex,parsep=1ex]
    \item We first justify the advantages of the StyleGAN architecture in disentanglement learning, by showing that StyleGAN augmented with a mutual information loss (called Info-StyleGAN) ourperforms most state-of-the-art unsupervised disentanglement methods.
    \item We propose Semi-StyleGAN that achieves near fully-supervised disentanglement quality with   limited supervision (0.25\%$\sim$2.5\%) on synthetic and real data.
    \item We propose new metrics (termed as MIG-gen and L2-gen) to evaluate the generator controllability, and reveal a crucial trade-off between learning disentangled representations and controllable generation.
    \item We then extend Semi-StyleGAN to an image-to-image model, enabling semantic fine-grained image editing with better generalization to unseen images.
    \item We create two high-quality datasets with much higher resolution, better photorealism,  and richer factors of variation than existing disentanglement datasets. 
\end{itemize}

\vspace{-2mm}
\section{Background and Related Work}

\paragraph{StyleGAN.}
GANs are a family of generative models that have shown great success. Among various GANs,
StyleGAN \citep{karras2019style} is a state-of-the-art GAN architecture for unsupervised image generation, particularly for high-fidelity human faces of resolution up to 1024x1024. 
StyleGAN comprises a mapping network whose role is to map a latent vector $z$ to an intermediate space, which then controls the styles at each convolutional layer in the synthesis network with adaptive instance normalization (AdaIN) \citep{ulyanov2016instance, huang2017arbitrary}.
StyleGAN also enables the separation of fine-grained and coarse-grained features. For example, modifying the styles of low-resolution blocks affects only coarse-grained features (e.g. overall pose and presence of eyeglasses), while modifying the styles of high-resolution blocks affects only fine-grained features (e.g. color scheme and microstructure). These nice properties make it a potentially good candidate for disentanglement learning of high-resolution images.



\vspace{-2mm}

\paragraph{Disentanglement learning.}

In terms of unsupervised disentanglement learning, there exists much prior work based on either Variational Autoencoders (VAEs) \citep{higgins2017beta,kim2018disentangling,chen2018isolating} or GANs \citep{chen2016infogan, lin2019infogan, nguyen2019hologan}. The basic idea in disentangled VAEs is to encourage a factorization of the latent code by regularizing the \textit{total correlation} \citep{chen2018isolating}.
They have shown the state-of-the-art performance on many standard disentanglement benchmarks \citep{dsprites17, kim2018disentangling}. 
Many GAN-based models reply on maximizing the mutual information between the observation and factor code, such as InfoGAN \citep{chen2016infogan} and its variants \citep{lin2019infogan}. 
Other GANs also learn disentanglement by designing a domain-specific generator architecture to add model-inductive bias, represented by HoloGAN \citep{nguyen2019hologan}. 
However, the use of 3D representations in HoloGAN may not scale up to higher-resolution images.

Another line of work in disentanglement learning is to use explicit supervision. \cite{kulkarni2015deep} applies a supervised training procedure to encourage each group of the graphics code to distinctly represent a specific factor of variation. \cite{bouchacourt2018multi} proposes the Multi-Level VAE (ML-VAE) to learn disentanglement from the supervision of group information. \cite{xiao2017dna} develops a supervised disentanglement algorithm called DNA-GAN using a swapping policy. 
\cite{narayanaswamy2017learning} proposes a semi-supervised VAE by employing a general graphical model structure in the encoder and decoder. However, it still remains unclear how the use of supervision impacts the disentanglement learning.
\cite{spurr2017guiding} proposes ss-InfoGAN by adding few labels to InfoGAN to learn semantically meaningful data representations.
More recently, \cite{locatello2019disentangling} shows the benefits of adding limited supervision into learning disentangled presentations in VAEs. We extend these results to GANs on more complex and higher-resolution images, and also quantify the impact of limited supervision on the generator controllability.

\paragraph{Conditional GANs.}

A class of conditional GANs conditions on the class labels for better image generation quality, such as cGAN \citep{mirza2014conditional}, AC-GAN \citep{odena2017conditional}, Projection Discriminator \citep{miyato2018cgans}.
The architectural component of semi-StyleGAN makes it a special case of semi-supervised conditional GANs, as it conditions on partially available factor codes. However, the tasks and loss functions of Semi-StyleGAN differ greatly from those of these conditional GANs. First, these conditional GANs mainly focus on  generating more realistic images, whereas semi-StyleGAN focuses on two joint tasks: (i) disentangled representation learning and (ii) controllable generation. Second, to this end, both the mutual information loss and a new smoothness regularization are introduced in our loss functions for a better trade-off in disentanglement learning.

Another class of conditional GANs conditions on the given images for image-to-image translation, where
many works focus on multi-attribute editing, and two representatives are StarGAN \citep{choi2018stargan} and AttGAN \citep{he2019attgan}. Although the proposed Semi-StyleGAN-\textit{fine} in Section \ref{sec_fine} share the similar objectives, there are many differences:
StarGAN needs to condition on images of a specific domain, namely a set of images sharing the same attribute, which limits itself to only performing discrete/binary attribute control. However, our work is capable of doing continuous style manipulation. AttGAN requires an encoder-decoder structure in the generator while we do not need to. Instead, our work adds controllable fine-grained factors along with a super-resolution process, which is especially suitable for editing fine styles of high-resolution images with a good generalization ability. 

\begin{figure}
  \centering
  \begin{subfigure}[b]{0.45\textwidth}
		\centering
		\small
		\includegraphics[width=\linewidth]{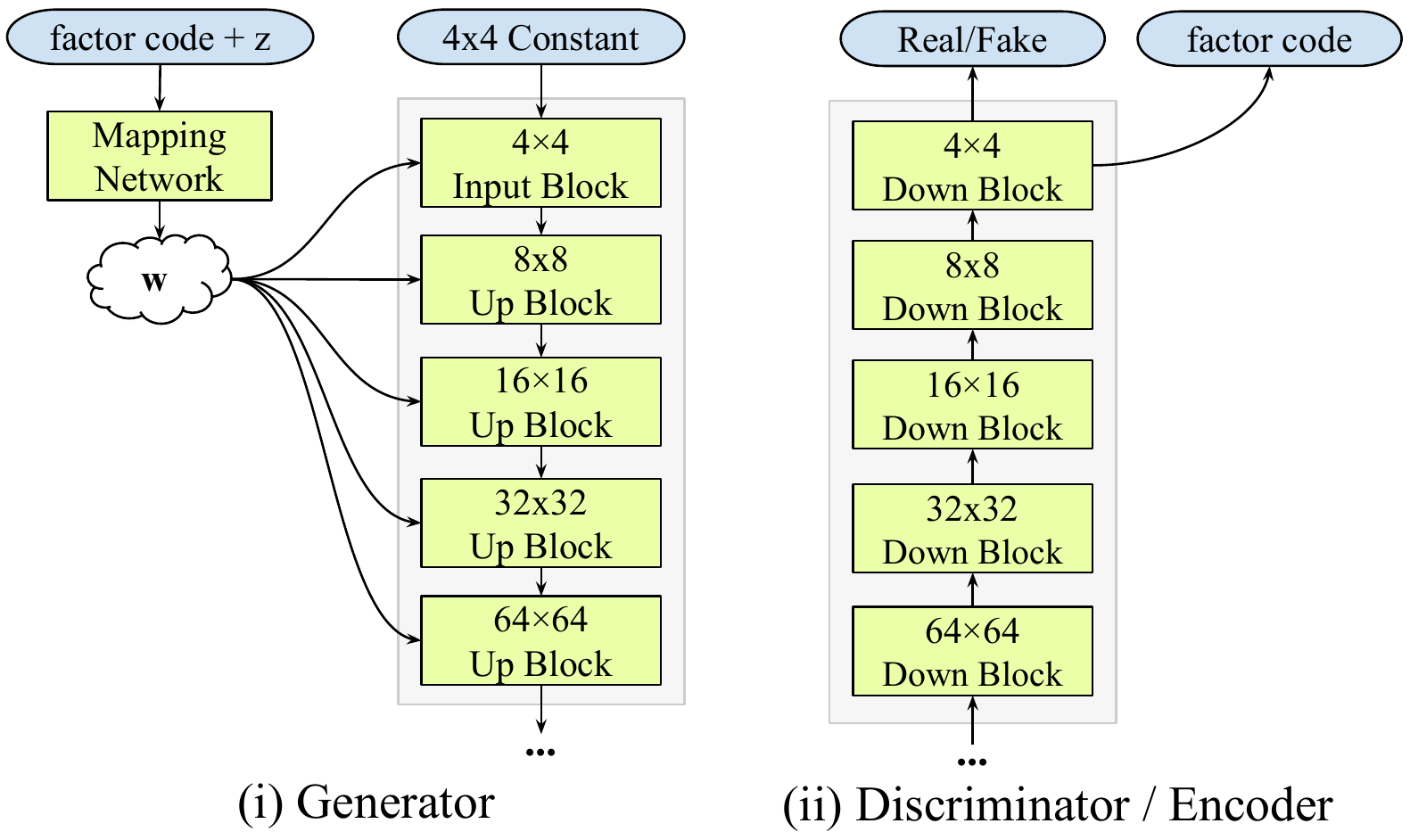}
  \end{subfigure}
  \vspace{-2mm}
  \caption{\small An illustration of disentanglement learning based on StyleGAN, where the mapping network in the generator conditions on the factor code and the encoder (which shares all layers in the discriminator except for the last layer) predicts its value.}
  \label{stylegan_dis}
\end{figure}

\vspace{-2mm}
\section{Why StyleGAN for Disentanglement?}

In unsupervised disentanglement learning, there exist many state-of-the-art VAE-based models, such as $\beta$-VAE \citep{higgins2017beta}, FactorVAE \citep{kim2018disentangling} and $\beta$-TCVAE \citep{chen2018isolating}. 
More recently, GAN-based models, such as InfoGAN-CR \citep{lin2019infogan}, have also achieved competitive performance by adding more tuning heuristics and regularization.
Here, we consider Info-StyleGAN, enabling StyleGAN with a mutual information loss, and show that the structural advance of StyleGAN provides a stronger prior for disentanglement learning compared to  regularization used previously in  VAEs or GANs.


As shown in Figure \ref{stylegan_dis}, the mapping network in the generator of Info-StyleGAN now conditions on a \textit{factor code}, a vector representing each factor of variation in each dimension, by simply concatenating it with the latent code $z$. The output of the mapping network, called \textit{conditional styles} will modulate each block in the synthesis network using AdaIN.
Similar to InfoGAN, the encoder in Info-StyleGAN shares all the network layers except the last one with the discriminator and predicts the value of factor code. Thus, we use $G / D / E$ to represent the generator, discriminator and encoder, respectively. The mutual information loss of InfoGAN can be approximated by an \textit{unsupervised code reconstruction} loss \citep{chen2016infogan}, which is
\begin{align}
    \mathcal{L}_{\text{unsup}} = \sum\nolimits_{c \sim \mathcal{C}, z \sim p_z } \left\Vert E(G(c, z)) - c \right\Vert_2
\end{align}
where $\mathcal{C}$ denotes the set of all factor codes, and $p_z$ denote the prior distribution of latent code $z$. The respective loss functions for $G$ and $(D, E)$ are given by
\begin{align} \label{loss_info}
\begin{split}
    \mathcal{L}^{(G)} = & \mathcal{L}_{\text{GAN}} + \gamma \mathcal{L}_{\text{unsup}} \\
    \mathcal{L}^{(D, E)} = & -\mathcal{L}_{\text{GAN}} + \gamma \mathcal{L}_{\text{unsup}}
\end{split}
\end{align}
where we keep the GAN loss function $\mathcal{L}_{\text{GAN}}$ the same as in \cite{karras2019style}. The hyperparameter $\gamma$ controls a trade-off between image realism and disentanglement quality.

\vspace{-2mm}
\subsection{Experimental Setup}
\vspace{-1mm}

\paragraph{Datasets and evaluation metrics. }
We consider two datasets to compare Info-StyleGAN and state-of-the-art disentanglement models: dSprites \citep{dsprites17} and our proposed Isaac3D (See details in Section \ref{sec_dataset}). dSprites is a commonly used dataset in disentanglement learning, with 737,280 images and each of resolution 64x64. For experiments on dSprites, we use both \textit{Factor score} \citep{kim2018disentangling} and \textit{Mutual Information Gap (MIG)} \citep{chen2018isolating} to evaluate disentanglement. For experiments on Isaac3D, we first downscale the resolution of each image to 128x128 because VAEs have difficulties in generating higher-resolution images. We use MIG to evaluate disentanglement quality and \textit{Frechet Inception Distance (FID)} \citep{heusel2017gans} to evaluate image quality.

\paragraph{Experimental protocol.}
We consider $\beta$-VAE, FactorVAE, $\beta$-TCVAE and InfoGAN-CR for comparison, where all the models are trained based on the implementation in \cite{locatello2019challenging}. 
For VAEs, we set $\beta=6$ for $\beta$-VAE, $\gamma=30$ for FactorVAE and $\beta=8$ for $\beta$-TCVAE after a grid search over different hyperparameters. 
For InfoGAN-CR, we use default hyperparameters on dSprites from the original paper, and perform a grid search over different hyperparameters on Isaac3D to report the best results.
For Info-StyleGAN, we keep $\gamma=10$ on dSprites and $\gamma=1$ on Isaac3D. We also keep the progressive training \citep{karras2017progressive}, as we show that it helps improve disentanglement in Appendix \ref{progressive}.
Because Info-StyleGAN and state-of-the-art disentanglement models have largely different network architectures, for a fairer comparison, we also try to keep their network sizes to be the same. 
See Appendix \ref{match_net_size} for how we decrease the network size of Info-StyleGAN (called Info-StyleGAN$^*$) to match those of previous models.

\begin{table}[t]
\begin{subtable}{\linewidth}
    \centering
    \footnotesize\addtolength{\tabcolsep}{-3pt}
    \begin{tabular}{c|c|c|c}
        \hline
         Methods & \# Params & Factor Score $\uparrow$ & MIG $\uparrow$ \\
         \hline
          $\beta$-VAE &
          0.69M &
0.713 $\pm$ 0.095 &
0.132 $\pm$ 0.031 \\
          FactorVAE & 
          5.70M &
0.764 $\pm$ 0.098 &
0.175 $\pm$ 0.057
\\
          $\beta$-TCVAE & 
          0.69M &
0.731 $\pm$ 0.097 &
0.174 $\pm$ 0.046
\\
InfoGAN-CR & 0.76M & \textbf{0.853 $\pm$ 0.046} & 0.270 $\pm$ 0.034
\\
\hline
        Info-StyleGAN* 
        & 
        0.74M &
0.769 $\pm$ 0.144 &
0.274 $\pm$ 0.096
\\
        Info-StyleGAN 
        & 47.89M &
{0.840 $\pm$ 0.090} & 
\textbf{0.290 $\pm$ 0.098}
\\
         \hline
    \end{tabular}
    \vspace{-2pt}
    \caption{dSprites with resolution 64x64}
\end{subtable} 

\vspace{2pt}

\begin{subtable}{\linewidth}
\centering
\footnotesize\addtolength{\tabcolsep}{-3pt}
\begin{tabular}{c|c|c|c}
        \hline
         Methods & \# Params & FID $\downarrow$ & MIG $\uparrow$ \\
         \hline
          $\beta$-VAE & 1.91M &
122.6 $\pm$ 2.0 &
0.231 $\pm$ 0.068 \\
          FactorVAE & 6.93M &
305.8 $\pm$ 142.1 &
0.245 $\pm$ 0.034
\\
          $\beta$-TCVAE & 1.91M &
155.4 $\pm$ 13.6 &
0.216 $\pm$ 0.074
\\
InfoGAN-CR & 3.29M & 80.72 $\pm$ 30.79 & 0.342 $\pm$ 0.139
\\
\hline
       Info-StyleGAN* 
        & 
        3.44M &
8.10 $\pm$ 2.25 &
\textbf{0.404 $\pm$ 0.085}
\\
       Info-StyleGAN 
        & 49.05M &
\textbf{2.19 $\pm$ 0.48} &
0.328 $\pm$ 0.057
\\
         \hline
    \end{tabular}
    \vspace{-2pt}
    \caption{Isaac3D with resolution 128x128}
    \end{subtable} 
    \vspace{-3mm}
    \caption{\small 
    Comparison of Info-StyleGAN and state-of-the-art disentanglement models on dSprites and (downscaled) Isaac3D. Note that the scores of VAEs are obtained based on the implementation in \cite{locatello2019challenging}. Info-StyleGAN* represents the smaller version of Info-StyleGAN, in which its number of parameters (i.e., \# Params) is similar to that of previous models.
 }
    \label{disent_score_vaes}
\end{table}

\vspace{-2mm}
\subsection{Key Results}
\vspace{-1mm}

Table \ref{disent_score_vaes} shows that Info-StyleGAN and its variant with smaller network size, termed as Info-StyleGAN$^*$, consistently outperform state-of-the-art VAE-based methods by a large margin on both dSprites and Isaac3D. Meanwhile, Info-StyleGAN achieves competitive or even better disentanglement performance than the strong GAN baseline.
Although unsupervised disentanglement learning is impossible without supervision or inductive bias \citep{locatello2019challenging}, this result reveals that the network structural improvement of StyleGAN provides \textit{a stronger prior} for disentanglement learning compared to different explicit loss regularizations in disentangled VAEs or InfoGAN-CR. Besides, we observe that previous methods have much higher FID scores 
on (downscaled) Isaac3D, along with their poor generated samples in Appendix \ref{vae_samples}. 
We have also increased the capacity of VAEs but the improvement of image quality still cannot close the gap with Info-StyleGAN, as shown in Appendix \ref{vae_samples_2}.
These results show that previous disentanglement methods have difficulties on more diverse and complex data, such as Isaac3D, while StyleGAN does not.  

\vspace{-2mm}
\section{Semi-StyleGAN}
\label{sec_semi_stylegan}
\vspace{-1mm}

As pointed out by \cite{locatello2019challenging} that unsupervised disentangled methods are formally non-identifiable \citep{hyvarinen1999nonlinear}, the impact of \textit{limited supervision} on both learning disentangled representations and controllable generation, which has been rarely explored, becomes crucial. In this section, we propose to add semi-supervision into Info-StyleGAN to get a semi-supervised disentanglement model -- \textit{Semi-StyleGAN}. Based on Semi-StyleGAN, we systematically analyze the role of limited supervision on both synthetic and real data.

A naive way of applying (semi-)supervision is to add a \textit{supervised code reconstruction} term for the small amount of labeled data into Eq (\ref{loss_info}), similar to \citep{kingma2014semi,odena2017conditional,locatello2019disentangling}. That is, 
\begin{align}
    \mathcal{L}_{\text{sup}} = \sum\nolimits_{(x, c) \sim \mathcal{J}}  \| E(x) - c \|_2
\end{align}
where $\mathcal{J}$ represents the set of labeled pairs of real image and factor code. When considering the \textit{limited supervision}, we assume the cardinality of the labeled set $\mathcal{J}$ satisfies that $|\mathcal{J}| \ll |\mathcal{X}|$, with $\mathcal{X}$ being the set of all real images.
Thus, the semi-supervised loss functions become
\begin{align} \label{loss_semi_naive}
\begin{split}
    \mathcal{L}^{(G)} = & \mathcal{L}_{\text{GAN}} + \gamma_G \mathcal{L}_{\text{unsup}} \\
    \mathcal{L}^{(D, E)} = & -\mathcal{L}_{\text{GAN}} + \gamma_E \mathcal{L}_{\text{unsup}} + \beta \mathcal{L}_{\text{sup}}
\end{split}
\end{align}
where $\beta$ is the weight of the supervised term $\mathcal{L}_{\text{sup}}$, and we use different $\gamma$'s (denoted by $\gamma_G$ and $\gamma_E$) to separately represent the weight of the unsupervised term in $\mathcal{L}^{(G)}$ and $\mathcal{L}^{(D, E)}$. As we show later, $\gamma_G$ and $\gamma_E$ play an important role in controlling the trade-off between encoder and generator disentanglement. Note that the supervised term $\mathcal{L}_{\text{sup}}$ does not update $G$ directly as shown in Eq. (\ref{loss_semi_naive}).

While semi-supervised learning for image recognition is an active research area, many algorithms may not be directly applied to disentanglement learning. Take the consistency regularization \citep{sajjadi2016regularization} as an example. Commonly used data perturbations, such as image rotation and color randomization will inevitably cause inconsistency if the considered factors of variation include object rotation or color. In contrast, encouraging smoothness in the latent space of GANs may help improve disentanglement \citep{karras2019style}. Thus, we propose to explicitly add a smoothness regularization by using the idea of MixUp \citep{zhang2018mixup, berthelot2019mixmatch}.

Formally, given a labeled observation-code pair $(x, c) \sim \mathcal{J}$ and a generated pair $(x', c')$ where $x' = G(z, c')$, 
we get a set of mixed observation-code pairs $\mathcal{M} = \left\{(\tilde{x}, \tilde{c})\right\}$ by
\begin{align} \label{mixup}
\begin{split}
    \lambda  \sim & \text{Beta} (\xi, \xi), \;\; \lambda' = \max (\lambda, 1 - \lambda) \\
    \tilde{x} &= \lambda' x + (1-\lambda') {x}' \\
    \tilde{c} &= \lambda' c + (1-\lambda') c' 
\end{split}
\end{align}
where $\xi$ is a hyperparameter. Thus, the smoothness regularization term is
\begin{align}
    \mathcal{L}_{\text{sr}} = \sum\nolimits_{(x, c) \sim \mathcal{M}}  \| E(x) - c \|_2
\end{align}
and the new semi-supervised loss functions with smoothness regularization become
\begin{align} \label{loss_semi_sr}
\begin{split}
    \mathcal{L}^{(G)} = & \mathcal{L}_{\text{GAN}} + \gamma_G \mathcal{L}_{\text{unsup}} + \alpha \mathcal{L}_{\text{sr}} \\
    \mathcal{L}^{(D, E)} = & -\mathcal{L}_{\text{GAN}} + \gamma_E \mathcal{L}_{\text{unsup}} + \beta \mathcal{L}_{\text{sup}} + \alpha \mathcal{L}_{\text{sr}}
\end{split}
\end{align}
where $\alpha$ is the weight of the smoothness term $\mathcal{L}_{\text{sr}}$.
Different from \cite{zhang2018mixup, berthelot2019mixmatch} that combines labeled and unlabeled real data, the MixUp in (\ref{mixup}) is performed between real labeled data and generated data. This way, it not only encourages smooth behaviors of both the generator and encoder, but also takes good advantages of enormous fake data for disentanglement.

\vspace{-2mm}
\subsection{New Datasets}
\label{sec_dataset}
\vspace{-1mm}

Current disentanglement datasets, such as dSprites \citep{dsprites17}, 3DShapes \citep{kim2018disentangling} and MPI3D \citep{gondal2019transfer}, are of low resolution and mostly lack photorealism. 
We create two new datasets -- \textit{Falcor3D} and \textit{Isaac3D}, with much higher resolution, better photorealism and richer factors of variations, as shown in Table \ref{dataset_desc}.

\vspace{-1mm}
\paragraph{Falcor3D.} 
It contains 233,280 images and each has a resolution of 1024x1024. This dataset is based on the 3D scene of a living room, where we move the camera positions and change the lighting conditions. 
Each image is paired with a ground-truth factor code, consisting of 7 factors of variation:
lighting intensity (5), lighting $x$-dir (6), lighting $y$-dir (6), lighting $z$-dir (6), camera $x$-pos (6), camera $y$-pos (6), and camera $z$-pos (6). 
The number $m$ behind each factor represents that the factor has $m$ possible values, uniformly sampled from $[0, 1]$. For example, ``lighting $x$-dir (6)'' represents the lighting direction moving along the $x$-axis and ``camera $z$-pos (6)'' denotes the camera position moving along the $z$-axis. Both factors have 6 possible values.

\vspace{-1mm}
\paragraph{Isaac3D.} 
It contains 737,280 images and each has a resolution of 512x512. This dataset is based on the 3D scene of a kitchen, where we move the camera positions and vary the lighting conditions. There is a robotic arm inside, grasping an object. 
The robotic arm has two degrees of freedom: $x$-movement (horizontal) and $y$-movement (vertical). The attached object could change its shape, scale or color.
All objects in the 3D scene are properly textured for better photorealism.
Similarly, each image is paired with a ground-truth factor code, consisting of 9 factors of variation: lighting intensity (4),  lighting $y$-dir (6),  object color (4), wall color (4), object shape (3), object scale (4), camera height (4), robot $x$-movement (8), and robot $y$-movement (5). The number $m$ behind each factor represents that it has $m$ possible values, uniformly sampled from $[0, 1]$. 


\begin{table}[t]
		\centering
		\small 
		\begin{tabular}{lcccc}
        \hline
         Datasets & \# Images & \# Factors & Resolution & 3D \\
         \hline
dSprites & 737,280 & 5 & 64x64 & \xmark\\
Noisy dSprites & 737,280 & 7 & 64x64 & \xmark\\
Scream dSprites & 737,280 & 7 & 64x64 & \xmark\\
SmallNORB & 48,600 & 5 & 128x128 & \cmark\\
Cars3D & 17,568 & 3 & 64x64 & \cmark\\
3DShapes & 480,000 & 7 & 64x64 & \cmark\\
MPI3D & 640,800 & 7 & 64x64 & \cmark\\
\textit{Falcor3D} & 233,280 & 7 & 1024x1024 & \cmark\\
\textit{Isaac3D} & 737,280 & 9 & 512x512 & \cmark\\
         \hline
    \end{tabular}
    \vspace{-3mm}
    \caption{\small Summary of the proposed two datasets, compared with currently commonly used datasets~\citep{gondal2019transfer}. We can see that the proposed two datasets -- \textit{Faclor3D} and \textit{Isaac3D} have much larger resolutions than previous datasets, along with the largest number of factors. More importantly, the proposed datasets are of much higher photorealism, as shown in Appendix \ref{datasets}.
    }
    \label{dataset_desc} 
\end{table}

\begin{figure*}
  \centering
  \begin{subfigure}[b]{0.47\textwidth}
		\centering
		\small
		\includegraphics[width=\linewidth]{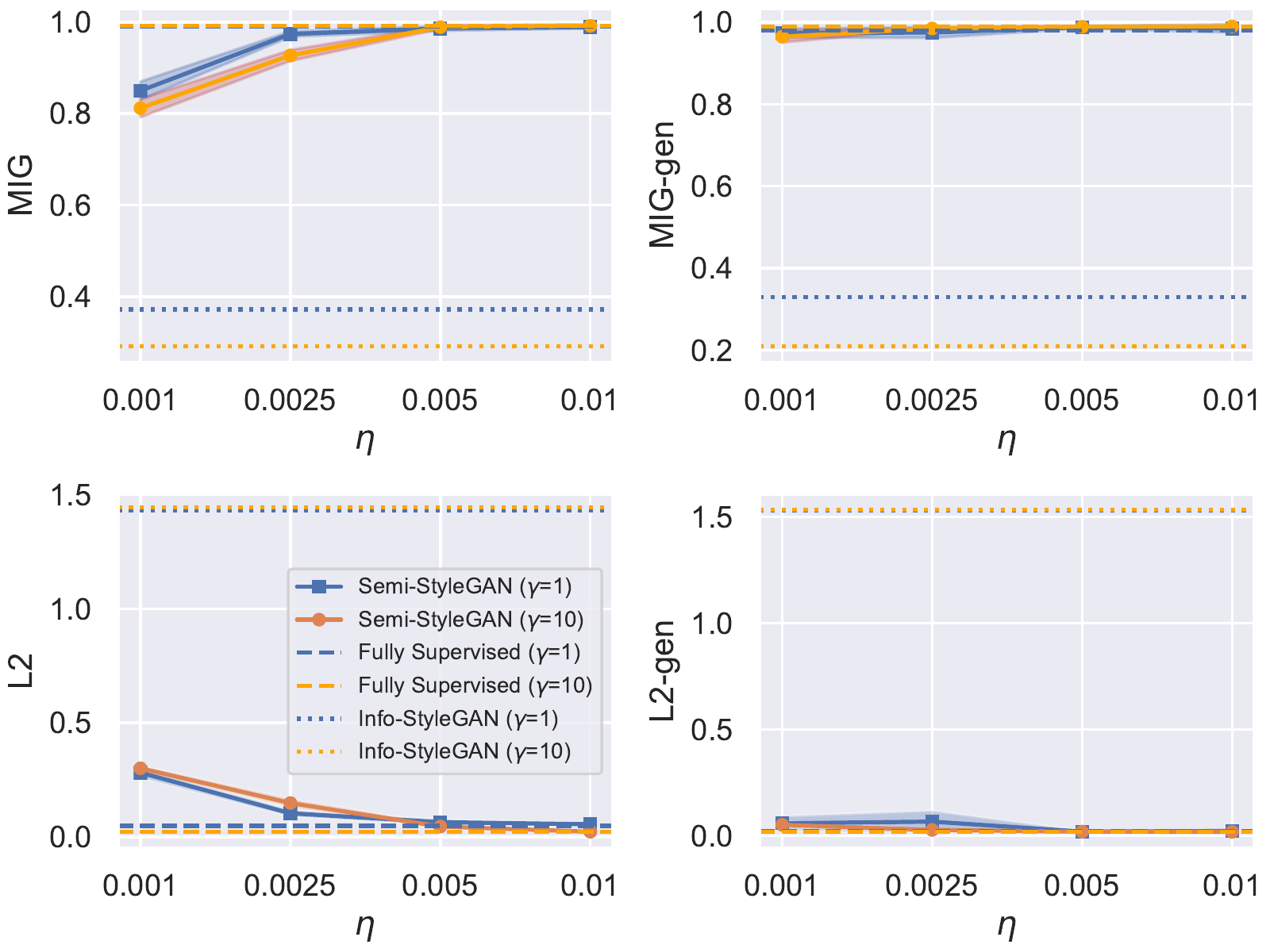}
		\vspace{-6mm}
		\caption{\small Isaac3D}
  \end{subfigure}
  \quad
  \begin{subfigure}[b]{0.47\textwidth}
		\centering
		\small
		\includegraphics[width=\linewidth]{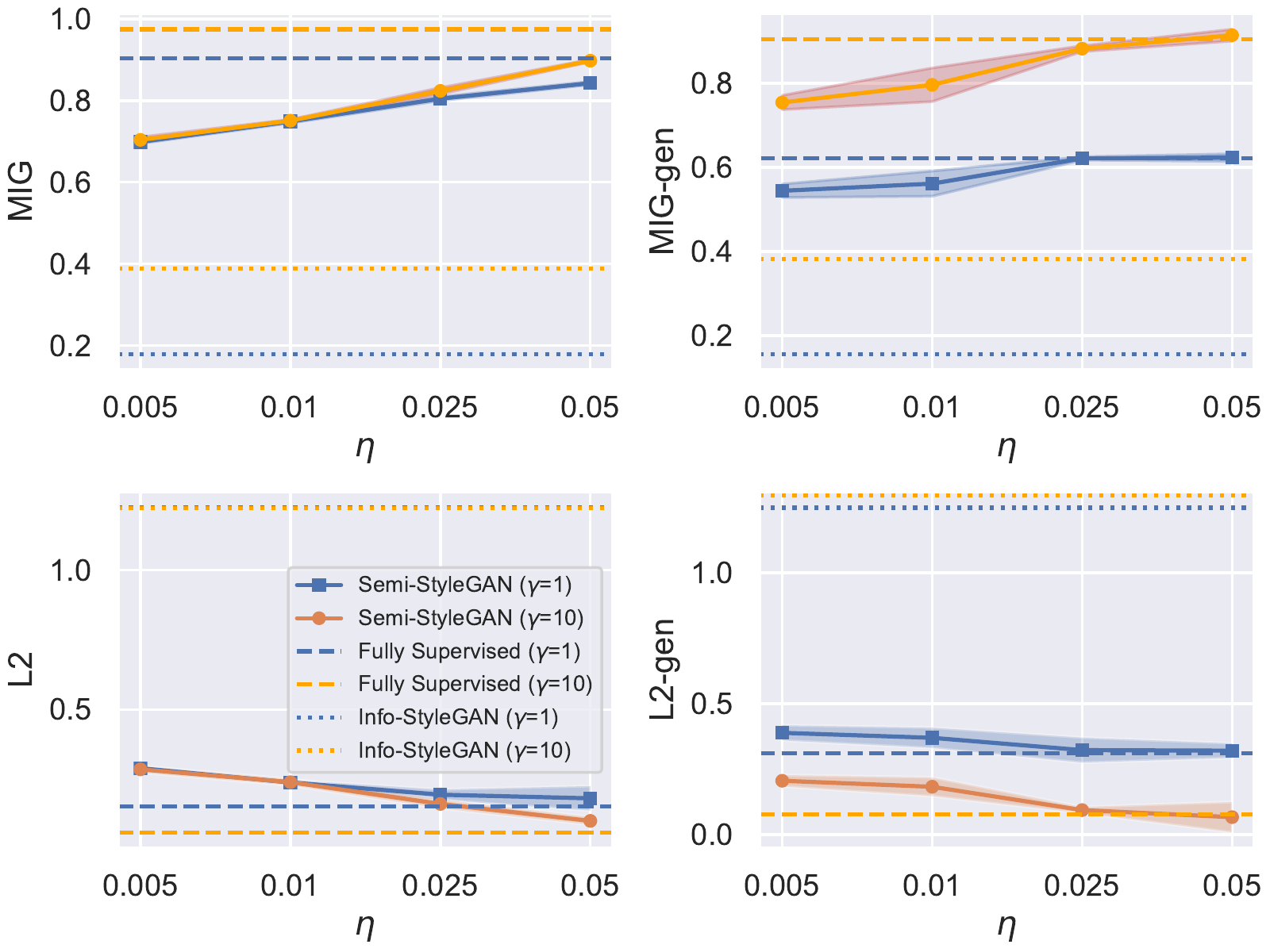}
		\vspace{-6mm}
		\caption{\small Falcor3D}
  \end{subfigure}
  \vspace{-3mm}
  \caption{\small Semi-StyleGAN with the default setting $\gamma_G=\beta=\gamma$, $\gamma_E=0$, $\alpha=1$ where $\gamma \in \{1, 10\}$ on (a) Isaac3D and (b) Falcor3D. We vary the portion of labeled data $\eta$ to show the impact of  semi-supervision by comparing with Info-StyleGAN (i.e. the unsupervised baseline), and the fully-supervised one ($\eta=1$).
  Only using 0.25$\sim$2.5\% of labeled data achieves near fully-supervised disentanglement.}
  \label{ss}
\end{figure*}

\begin{table*}[t]
\begin{subtable}{\linewidth}
\centering
\footnotesize\addtolength{\tabcolsep}{-2pt}
\begin{tabular}{c|c|c||c|c}
        \hline
          {Methods} & MIG $\uparrow$ & L2 $\downarrow$ & MIG-gen $\uparrow$ & L2-gen $\downarrow$ \\
         \hline
         Encoder-only & 0.731 $\pm$ 0.009 & {0.379 $\pm$ 0.002} & - & - \\
         Encoder-only w/ MixUp & 0.834 $\pm$ 0.004 & {0.279 $\pm$ 0.005} & - & - \\
         \hline
         Semi-StyleGAN & 0.812 $\pm$ 0.020 & 0.301 $\pm$ 0.012 & \textbf{0.965 $\pm$ 0.014} & \textbf{0.052 $\pm$ 0.016} \\
         + Remove smoothness consistency & 0.765 $\pm$ 0.042 & 0.347 $\pm$ 0.019 & {0.945 $\pm$ 0.011} & {0.072 $\pm$ 0.008} \\
         + Increase the $\mathcal{L}_{\text{unsup}}$ term in $E$ ($\gamma_E=10$) & \textbf{0.880 $\pm$ 0.120} & \textbf{0.225 $\pm$ 0.222} & 0.888 $\pm$ 0.087 & 0.283 $\pm$ 0.247 \\
         + Remove the $\mathcal{L}_{\text{unsup}}$ term in $G$ & {0.719 $\pm$ 0.014} & {0.490 $\pm$ 0.024} & 0.130 $\pm$ 0.054 & 1.514 $\pm$ 0.003 \\
    \hline
    \end{tabular}
    \vspace{-1mm}
    \caption{Isaac3D ($\eta$ = 0.1\%)}
\end{subtable} 
\vspace{2pt}
\begin{subtable}{\linewidth}
\centering
\footnotesize\addtolength{\tabcolsep}{-2pt}
\begin{tabular}{c|c|c||c|c}
        \hline
          {Methods} & MIG $\uparrow$ & L2 $\downarrow$ & MIG-gen $\uparrow$ & L2-gen $\downarrow$ \\
          \hline
         Encoder-only & {0.690 $\pm$ 0.007} & {0.271 $\pm$ 0.002} & - & - \\
         Encoder-only w/ MixUp & 0.701 $\pm$ 0.005 & \textbf{0.265 $\pm$ 0.003} & - & - \\
         \hline
          Semi-StyleGAN & \textbf{0.704 $\pm$ 0.007} & 0.285 $\pm$ 0.002 & \textbf{0.754 $\pm$ 0.017} & \textbf{0.205 $\pm$ 0.022}
         \\
         + Remove smoothness consistency & {0.674 $\pm$ 0.011} & {0.296 $\pm$ 0.017} &  {0.632 $\pm$ 0.058} & {0.303 $\pm$ 0.088}
         \\
         + Increase the $\mathcal{L}_{\text{unsup}}$ term in $E$ ($\gamma_E=10$) & 0.643 $\pm$ 0.035 & 0.343 $\pm$ 0.016 & 0.636 $\pm$ 0.065 & 0.346 $\pm$ 0.070 \\
         + Remove the $\mathcal{L}_{\text{unsup}}$ term in $G$ & {0.680 $\pm$ 0.016} & {0.300 $\pm$ 0.010} & 0.034 $\pm$ 0.028 & 1.096 $\pm$ 0.086 \\
         \hline
    \end{tabular}
    \vspace{-1mm}
    \caption{Falcor3D ($\eta$ = 0.5\%)}
    \end{subtable} 
    \vspace{-5mm}
    \caption{\small 
    Ablation studies of Semi-StyleGAN on (a) Isaac3D and (b) Falcor3D, where the default setting is $\gamma_G=\beta=10$, $\gamma_E=0$, $\alpha=1$. 
    ``Encoder-only'' means we train the encoder by minimizing the L2 score with the labeled data only, a supervised baseline for the encoder disentanglement. ``Encoder-only w/ MixUp'' means we train the encoder by using MixUp \citep{zhang2018mixup}, a semi-supervised baseline for the encoder disentanglement.
    We set $\eta=0.1\%$ on Isaac3D and $\eta=0.5\%$ on Falcor3D, respectively. 
 }
    \label{ss_ablation}
\end{table*}

\vspace{-2mm}
\subsection{New Metrics}

Many metrics have been proposed for evaluating disentanglement, such as
Factor score \citep{kim2018disentangling}, MIG \citep{chen2018isolating},
DCI score \citep{ridgeway2018learning},
and SAP score \citep{kumar2017variational}. See the prior work \citep{locatello2019challenging} for their more implementation details. 
However, they all have some inherent limitations in quantifying semi-supervised disentanglement methods. 

First, these metrics are designed for \textit{unsupervised} disentanglement methods, which are non-identifiable \citep{locatello2019challenging}. But with supervision, the model become identifiable and thus we need to evaluate the semantic meaning of learned representations as well. 
A simple solution here is to measure the average $L2$ distance between the ground-truth factor code and the prediction of its paired observation using the considered encoder, termed as $L2$ score.

Second, these metrics only evaluate the the encoder disentanglement while ignoring the generator controllability, another important characteristic of disentanglement learning. 
However, there may exist a trade-off between the encoder and generator disentanglement. That is, a high MIG score does not mean a good model in terms of the controllable generation ability. Therefore, we propose new metrics to quantify the generator controllability.


Specifically, given a generator $G$ to be evaluated and an \text{oracle encoder} $E_{\text{oracle}}$ that can perfectly predict the factor code, we first sample $N$ generated observation-code pairs $( {x'}^{(n)}, {c'}^{(n)} )$ where ${x'}^{(n)} = G(z, {c'}^{(n)})$. We then pass the generated sample ${x'}^{(n)}$ into $E_{\text{oracle}}$ to get its factor code prediction $\hat{c}'^{(n)} = E_{\text{oracle}}(x'^{(n)})$. Accordingly, we measure the correlation between $\hat{c}'^{(n)}$ and ${c'}^{(n)}$ in the same way with prior disentanglement metrics. In particular, we define an MIG-like metric, called \textit{MIG-gen}, to evaluate the generator,
\begin{align*}
\text{MIG-gen} = \frac{1}{NK} & \sum\limits_{n=0}^{N-1}\sum\limits_{k=0}^{K-1} \frac{1}{H({\hat{c}^{(n)}_k})} \cdot \\ 
& \left( I(\hat{c}'^{(n)}_{j_k}; {c'}^{(n)}_k) - \max\limits_{j \neq j_k} I(\hat{c}'^{(n)}_{j}; {c'}^{(n)}_k) \right)
\end{align*}
where $K$ is the length of factor code, $H(\cdot)$ and $I(\cdot;\cdot)$ denote the entropy and mutual information, respectively, and $j_{k} = \arg\max\nolimits_{j} I(\hat{c}'^{(n)}_j, {c'}^{(n)}_k)$. 
Similarly, we also introduce \textit{L2-gen} to measure the semantic correctness of the generator,
$$ \text{L2-gen} = \frac{1}{N} \sum\limits_{n=0}^{N-1} \| E_{\text{oracle}}({x}'^{(n)}) - {c'}^{(n)} \|_2 $$
Intuitively, if the oracle encoder is perfect for every ground-truth observation-code pair, 
any mismatch between its prediction and the corresponding factor code should be contributed to the generator instead.
Thus, both MIG-gen and L2-gen can effectively measure the generator controllability.
To obtain an oracle encoder for each dataset, such as the proposed Falcor3D and Isaac3D, we pre-train a separate encoder network by minimizing the $L2$ score with all the ground-truth observation-code pairs. 

\vspace{-1mm}
\subsection{Experimental Setup} 
\vspace{-1mm}

\paragraph{Datasets and evaluation metrics.}
To test the proposed Semi-StyleGAN on complex high-resolution images 
that many prior works have difficulty with, 
we focus on three datasets: Isaac3D with resolution 512x512, Falcor3D with resolution 512x512 and CelebA with resolution 256x256. 
For the proposed Isaac3D and Falcor3D, we use \textit{MIG} and \textit{L2} to measure the encoder disentanglement, and \textit{MIG-gen} and \textit{L2-gen} to measure the generator controllability. For experiments on CelebA, we focus on the latent traversals to qualitatively measure the disentanglement quality.

\vspace{-1mm}
\paragraph{Experimental protocol.}
Before training, we first get the labeled set $\mathcal{J}$, by randomly sampling observation-code pairs from each dataset with a probability $\eta$. All the remaining observations form as the unlabeled set. The value of $\eta \in [0, 1]$ controls the portion of labeled data during training. Particularly Semi-StyleGAN becomes a fully-supervised method if $\eta = 1$, and reduces to Info-StyleGAN if $\eta = 0$. 
In experiments, we set $\xi=0.75$ in Eq. (\ref{mixup}) to be the same with \cite{berthelot2019mixmatch}. For the hyperparameters $\{\gamma_G, \gamma_E, \beta, \alpha\}$, we find that setting $\gamma_G=\beta=\gamma$, $\gamma_E=0$, $\alpha=1$ works well across different datasets, where we vary $\gamma \in \{1, 10\}$.
Thus, without stated otherwise, we use the above setting by default in Semi-StyleGAN.

Our experiments mainly include four aspects. (i) We first vary the supervision rate $\eta$ to show the impact of limited supervision. 
(ii) Given the supervision rate $\eta$, we train the encoder alone with the labeled data only (called \textit{Encoder-only}) and with the MixUp (called \textit{Encoder-only w/ MixUp}), respectively, as supervised and semi-supervised baselines for the encoder disentanglement.
(iii) For ablation studies, we vary $\gamma_G, \gamma_E$ to reveal the trade-off between the encoder and generator disentanglement, and vary $\alpha$ to show the impact of smoothness regularization.
(iv) We show latent traversal results on both synthetic and real datasets.


\vspace{-2mm}
\subsection{Key Results}
\label{semi_key_results}
\vspace{-1mm}

\paragraph{Impact of limited supervision.} Figure \ref{ss} shows the quantitative results of varying $\eta$ in Semi-StyleGAN on Isaac3D and Falcor3D, where we consider two cases of $\gamma \in \{1, 10\}$. 
For Isaac3D, only using 0.25\% of the labeled data
can achieve a very close performance to the fully-supervised one ($\eta=1$) in terms of both the encoder and generator disentanglement.
Similarly for Falcor3D, only using 2.5\% of the labeled data 
can also achieve near the fully-supervised disentanglement.
It means adding a very small amount of labeled data (0.25\%$\sim$2.5\%) into the training dataset could benefit significantly the disentanglement learning with Semi-StyleGAN.
Besides, we can see that the generator disentanglement is more sensitive to the choice of $\gamma$ than the encoder disentanglement, particularly on Falcor3D.

\begin{figure} [t]
  \centering
  \begin{subfigure}[b]{0.49\textwidth}
		\centering
		\includegraphics[width=\linewidth]{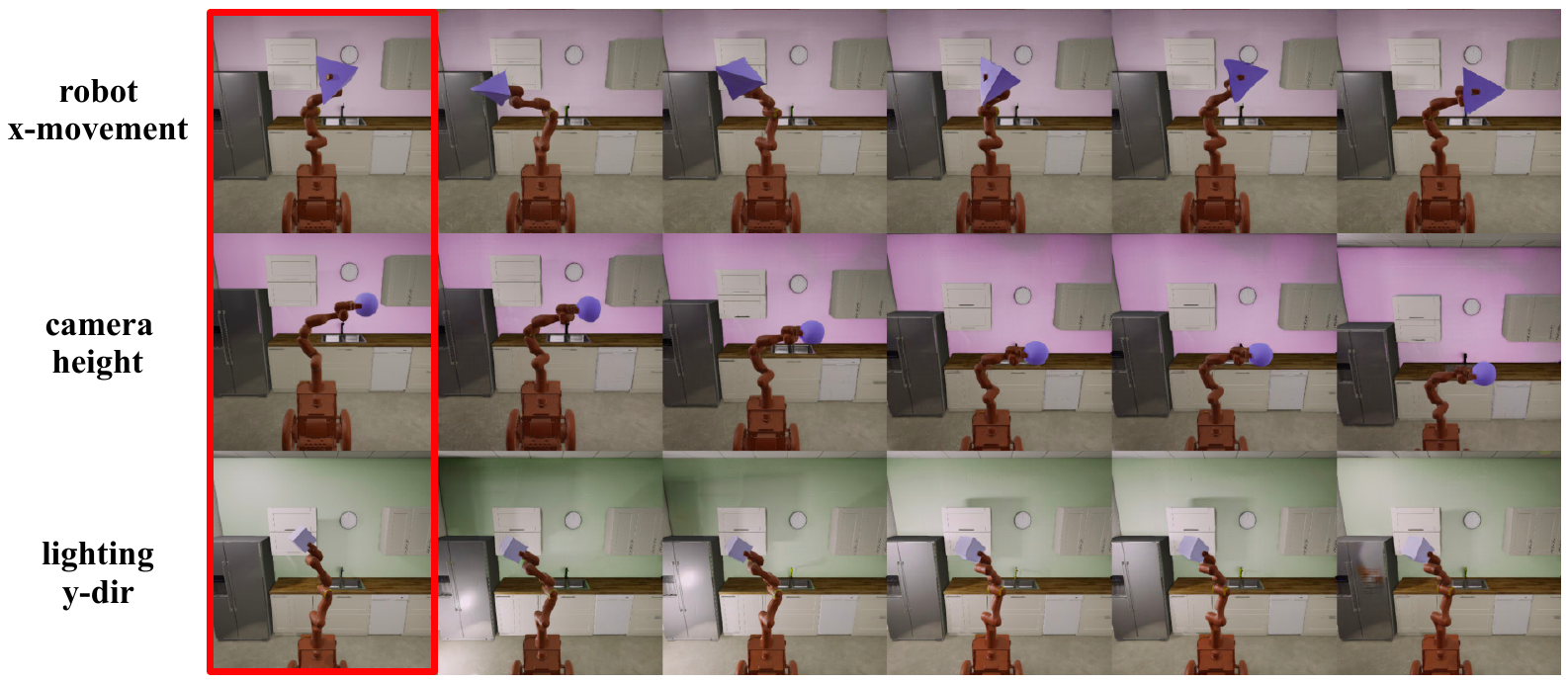}
		\caption{\small Semi-StyleGAN on Isaac3D with 0.5\% of labeled data}
  \end{subfigure}
  \begin{subfigure}[b]{0.49\textwidth}
		\centering
		\includegraphics[width=\linewidth]{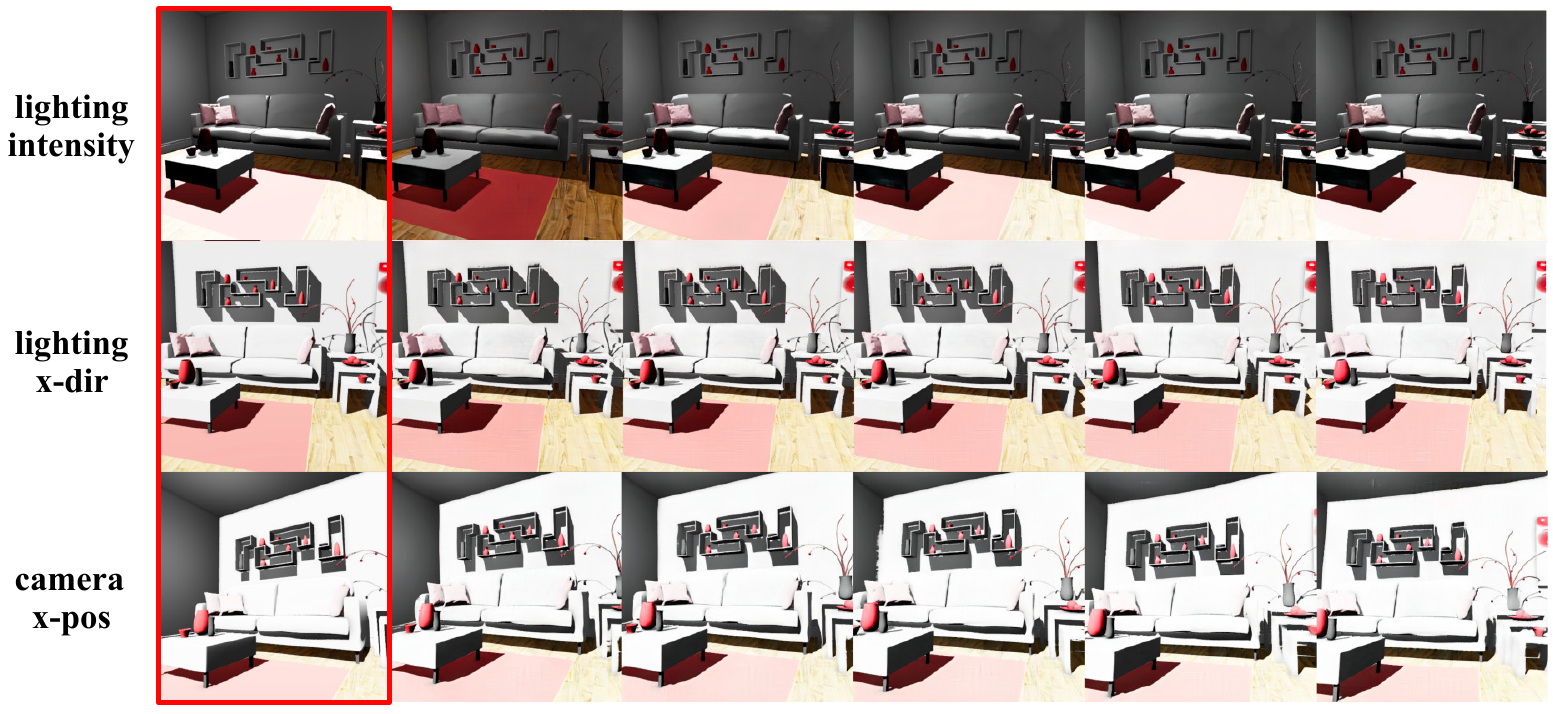}
		\caption{\small Semi-StyleGAN on Falcor3D with 1\% of labeled data}
  \end{subfigure}
  \vspace{-6mm}
  \caption{\small Latent traversal of Semi-StyleGAN on Isaac3D and Falcor3D where $\gamma =10$. Images in the first column (marked by red box) are randomly sampled real images and the rest images in each row are their interpolations, by uniformly varying the given factor from 0 to 1. See Appendix \ref{sec_semi_isaac3d} and \ref{sec_semi_falcor3d} for more results.
} 
  \label{latent_trav_synt}
\end{figure}

\vspace{-2mm}
\paragraph{Ablation studies and comparison with baselines. } First, Table \ref{ss_ablation} shows there may exist a crucial trade-off between the encoder and generator disentanglement, governed by the interplay between the unsupervised and supervised loss terms. For example, 
we can see that Semi-StyleGAN gets the best generator disentanglement by slightly sacrificing the encoder disentanglement, in particular on Isaac3D.
Second, by removing the smoothness consistency, we can see a large performance drop in terms of all metrics on both datasets,  which demonstrates the effectiveness of the smoothness regularization in Semi-StyleGAN.
Besides, there exist a trade-off between the generator controllability and encoder disentanglement. For example, increasing the $\mathcal{L}_{\text{unsup}}$ term in $E$ ($\gamma_E=10$) worsens the generator controllability while improving the encoder disentanglement on Isaac3D.
Removing the $\mathcal{L}_{\text{unsup}}$ term in $G$ ($\gamma_G = 0$), we can still get a decent encoder disentanglement despite the generator controllability totally fails. This trade-off also depends on the datasets, evidenced by  different behaviors on Isaac3D and Falcor3D. 
Finally, Table \ref{ss_ablation} shows the results of comparing Semi-StyleGAN with the supervised and semi-supervised baselines. We can see that Semi-StyleGAN significantly outperforms the supervised baseline (i.e., Encoder-only) on Isaac3D, and also gets a better MIG score on Falcor3D. 
With good hyperparameters which potentially weigh more on the encoder side, we can also achieve on par or better encoder disentanglement than the semi-supervised baseline (i.e., Encoder-only w/ MixUp). 



\begin{figure} [t]
  \centering
  \begin{subfigure}[b]{0.49\textwidth}
		\centering
		\includegraphics[width=\linewidth]{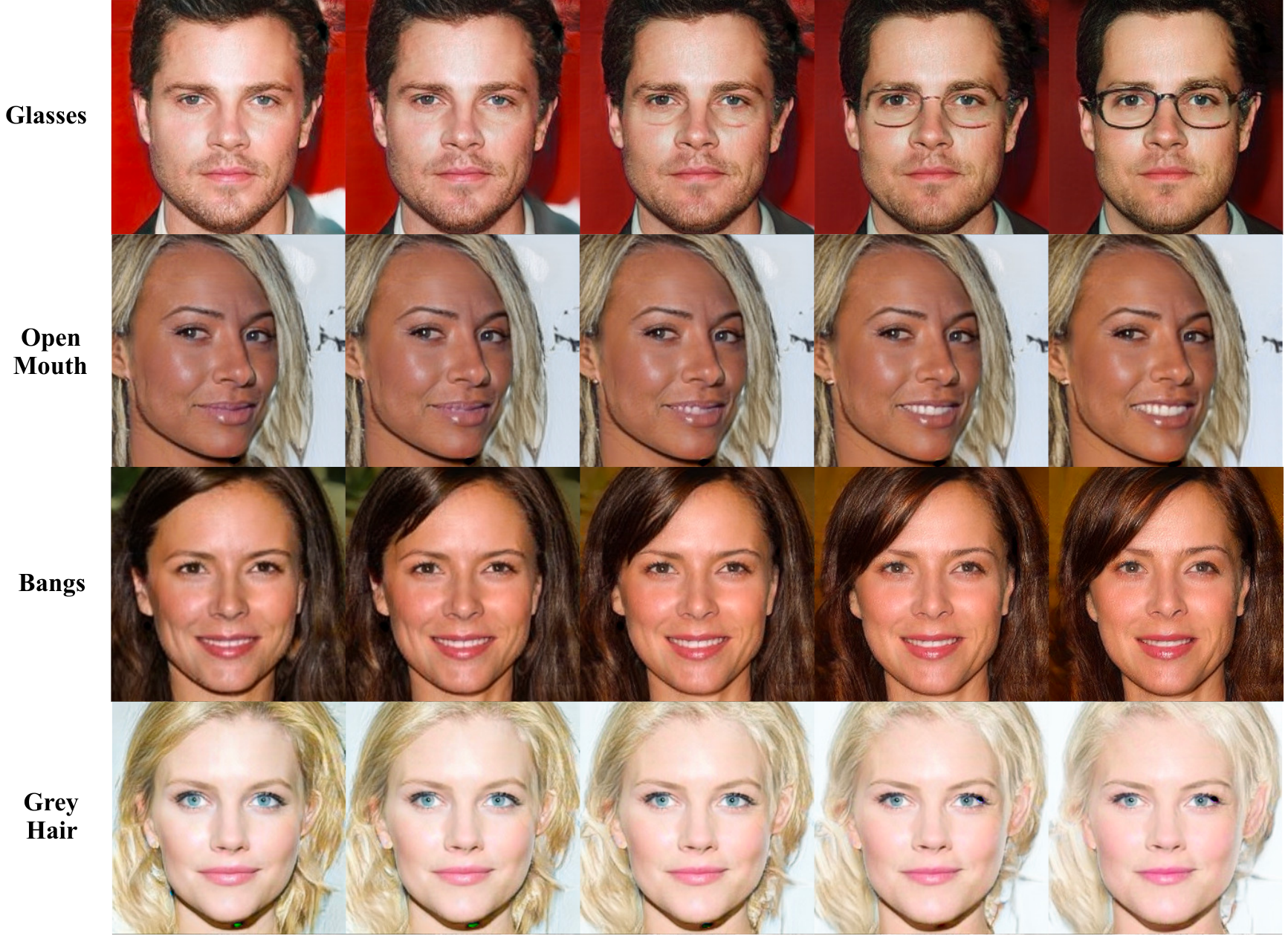}
  \end{subfigure}
  \vspace{-6mm}
  \caption{\small Latent traversal of Semi-StyleGAN on CelebA with resolution 256x256 by using 0.5\% of the labeled data, where we use $\gamma=1$ and disentangle all 40 binary attributes. See Appendix \ref{sec_semi_celeba} for the results of other attributes.
} 
  \label{latent_trav_celeba}
\end{figure}

\vspace{-2mm}
\paragraph{Latent traversal on synthetic and real data. } Qualitatively, we show the latent traversal results on both synthetic and real data in Figure \ref{latent_trav_synt} and \ref{latent_trav_celeba}.
When only 0.5\% or 1\% of the labeled data is available, each factor in the interpolated images changes smoothly without affecting other factors. For Isaac3D and Falcor3D, all the interpolated images visually look the same with their reference real image except for the considered factor, verifying the semantic correctness of Semi-StyleGAN with very limited supervision. 
For CelebA, we use a higher resolution than the prior work \citep{chen2016infogan, higgins2017beta, nguyen2019hologan}, and achieve visually better disentanglement quality together with a higher image quality. It means that the insights gained in the synthetic datasets also applies to the real domain. With very limited supervision, Semi-StyleGAN can achieve good disentanglement on real data.

\vspace{-2mm}
\section{An Extension for Better Generalization} \label{sec_fine}
\vspace{-1mm}

Although Semi-StyleGAN performs well in both synthetic and real data, it cannot generalize to unseen data whose high-level content does not match the training data but whose fine-grained styles might. For instance, Semi-StyleGAN trained on Isaac3D cannot generate an image in which the robot arm stands on the right (instead of in the middle as in the training data).
In this section, we design a new GAN architecture that extends Semi-StyleGAN to an image-to-image model, that we call Semi-StyleGAN-\textit{fine}. This model achieves better generalization to unseen data. 

\begin{figure}
  \centering
  \begin{subfigure}[b]{0.45\textwidth}
		\centering
		\small
		\includegraphics[width=\linewidth]{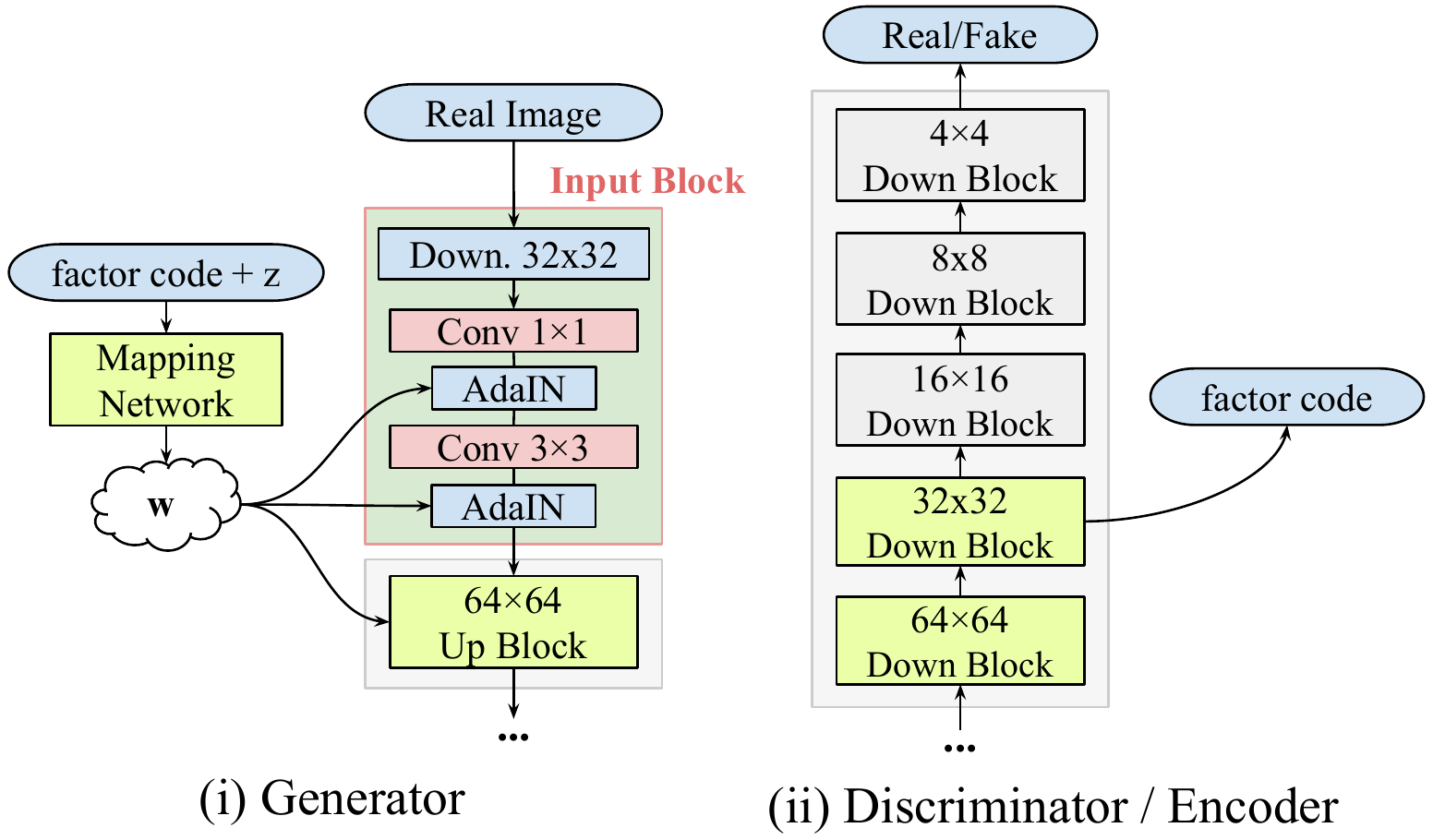}
  \end{subfigure}
  \vspace{-3mm}
  \caption{\small An illustration of Semi-StyleGAN-\textit{fine}, where we downsample the real image into 32x32 resolution and replace the lower resolution blocks (4x4 - 32x32) in the generator by a new input block. Also, the encoder predicts the value of (fine-grained) factor code from the 32x32 block instead.}
  \label{semi-stylegan-ext}
\end{figure}

Inspired by the observations that lower resolution blocks in the StyleGAN generator learn coarse-grained features while its high-resolution blocks account for fine-grained styles, we change the StyleGAN generator to not contain lower-resolution blocks. As shown in Figure \ref{semi-stylegan-ext}, the generator instead takes the real image as one of its inputs by downscaling it to a lower resolution $\phi$ (e.g., $\phi = 32$ in Figure \ref{semi-stylegan-ext}). Accordingly, it generates the high-resolution image by only modulating the (fine-grained) factor code into higher resolution blocks.
Also, the encoder predicts the value of factor code from the block with resolution $\phi$, instead of the last output block. 
The intuition is that 
lower-resolution blocks in the encoder also have less relationship with fine-grained styles, and thus the code prediction better use its higher resolution blocks only. 
This way, the generator in Semi-StyleGAN-\textit{fine} does a semi-supervised controllable 
fine-grained image editing while its encoder infers the fine-grained factor code that the generator has used. Finally, the loss functions remain the same as in Eq. (\ref{loss_semi_sr}).


\vspace{-2mm}
\subsection{Experimental Setup}
\vspace{-1mm}

We mainly focus on the latent traversals of Semi-StyleGAN-\textit{fine} on Isaac3D and CelebA to qualitatively test its generalization ability. 
In the training time, we train Semi-StyleGAN-\textit{fine} on Isaac3D and CelebA, respectively. In the test time, we apply novel test images (with the different high-level content) as the input to evaluate the proposed method.
For Isaac3D, the novel test images are given by: (i) shifting the robot position to the right side (instead of standing right in the middle for all the training data), and (ii) attaching the robot arm with an unseen object. For CelebA, we simply download some new face images from online, following by aligning and cropping them into 256x256.

\begin{figure} [t]
  \centering
  \small
  \begin{subfigure}[b]{0.48\textwidth}
		\centering
		\includegraphics[width=\linewidth]{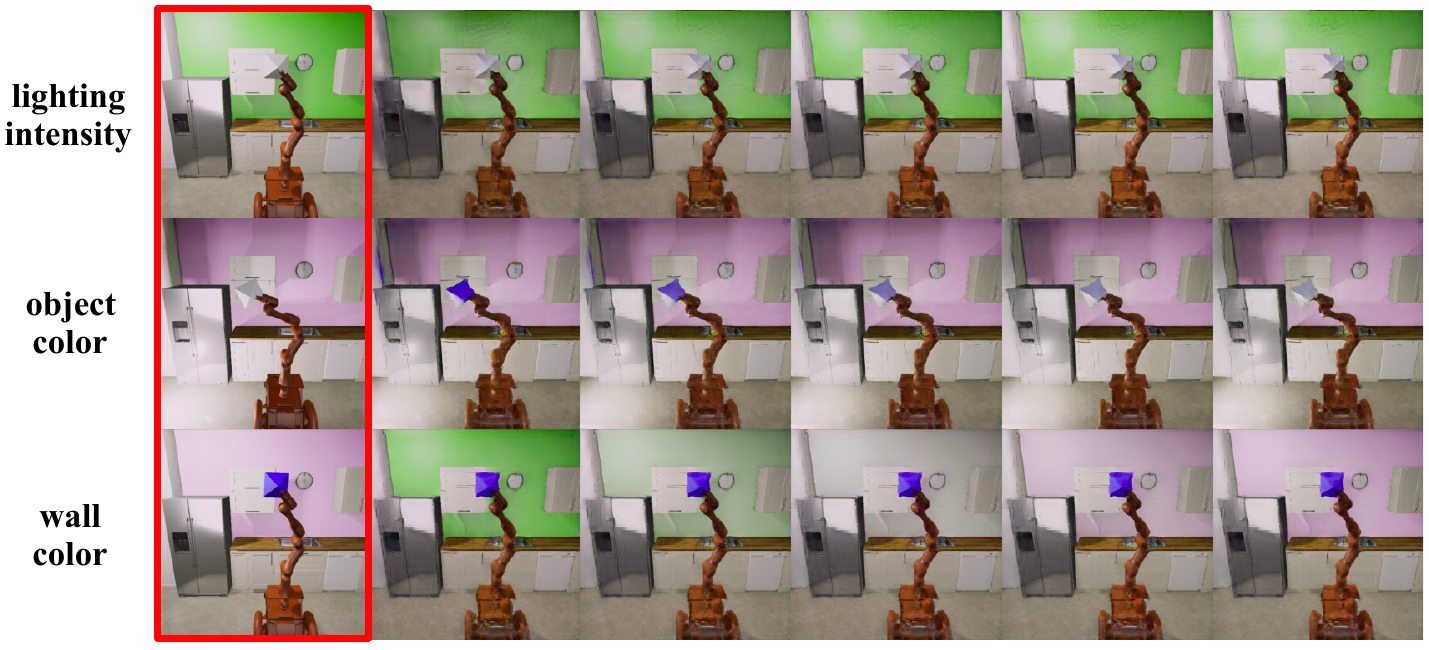}
  \end{subfigure}
  \vspace{-6mm}
  \caption{\small Generalized latent traversal results of Semi-StyleGAN-\textit{fine} trained on Isaac3D with 1\% of labeled data where we set $\phi=64$ and interpolate the shown fine styles. In the test images, we shift the position of the robot arm to the right side, and also attach it with an unseen object (i.e., a tetrahedron).    
  } 
  \label{ext_isaac3d}
\end{figure}

\begin{figure} [t]
\setlength\belowcaptionskip{-10pt}
  \centering
  \small
  \begin{subfigure}[b]{0.48\textwidth}
		\centering
		\includegraphics[width=\linewidth]{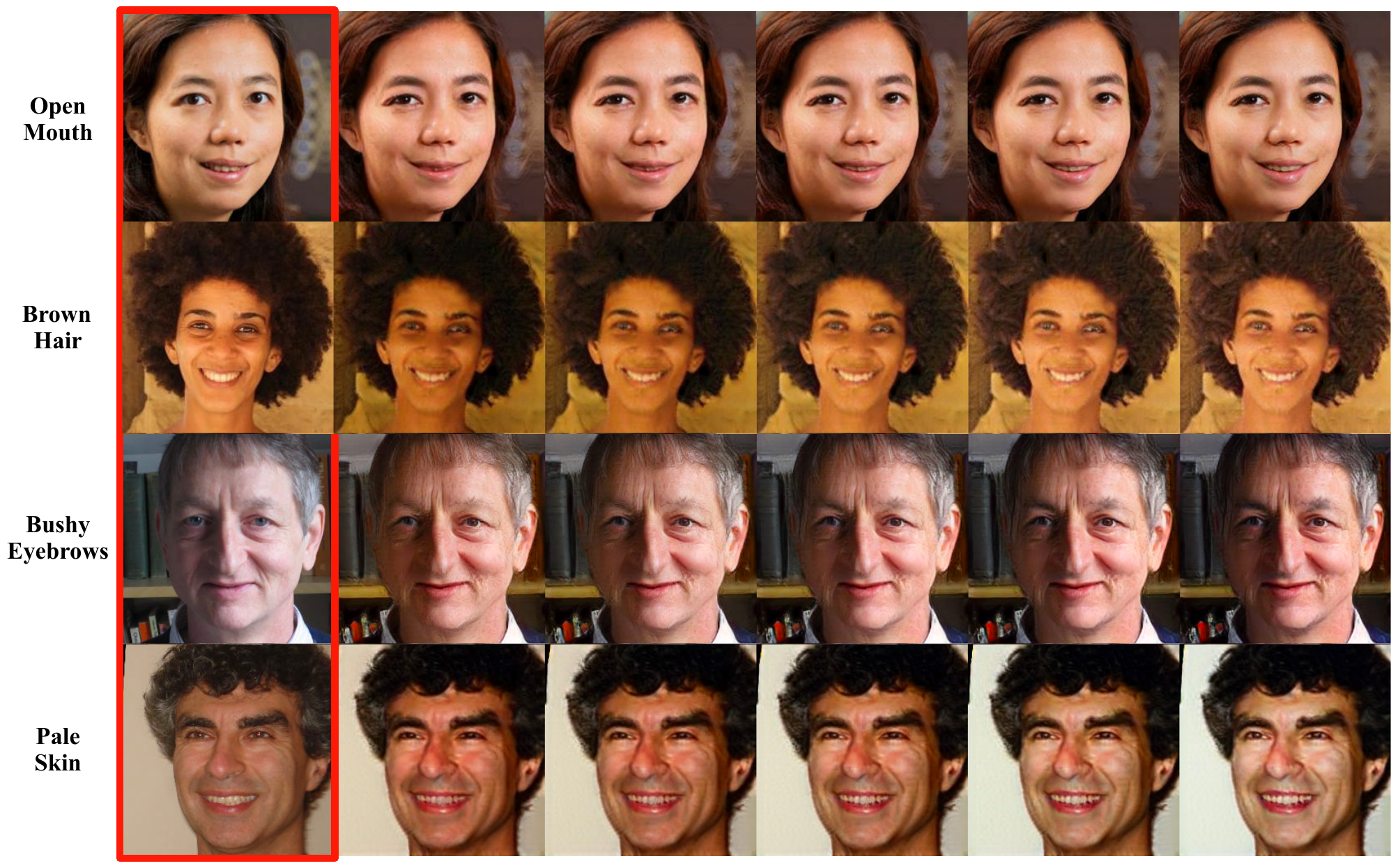}
  \end{subfigure}
  \vspace{-6mm}
  \caption{\small Generalized latent traversal results of Semi-StyleGAN-\textit{fine} trained on CelebA with 1\% of labeled data where we set $\phi=64$ and control the shown fine styles. 
  } 
  \label{ext_celeba}
  \vspace{-4mm}
\end{figure}

\vspace{-2mm}
\subsection{Key Results}
\vspace{-1mm}
Figure \ref{ext_isaac3d} and \ref{ext_celeba} show the results of interpolating fine-grained factors in the Isaac3D dataset and the CelebA dataset, respectively, with different test images where we set $\eta=0.01$ and $\phi=64$. We can see that each considered fine-grained factor in both datasets keeps changing smoothly during its interpolations without affecting other factors, implying good generalized disentanglement. 
Particularly, the interpolations of the Isaac3D test images all maintain the new robot position and new object shape (i.e., a tetrahedron) with relatively high image quality. See Appendix \ref{sec_semi_ext} for test images with another novel object shape. The interpolations of the CelebA test images also keep the same identities and other coarse-grained features with the given input images. 
It is worthy to note that the good generalized disentanglement of Semi-StyleGAN-\textit{fine} has been achieved by using only 1\% of labeled data, and further increasing the supervision rate $\eta$ does not visually improve performance.
Therefore, these results demonstrate the ability of Semi-StyleGAN-\textit{fine} in semantic fine-grained image editing with limited supervision that generalizes well to unseen novel images.


\vspace{-2mm}
\section{Conclusions}
\vspace{-1mm}

In this paper, we designed new loss functions and architectures based on StyleGAN for semi-supervised high-resolution disentanglement learning. 
We first showed that Info-StyleGAN largely outperforms most state-of-the-art unsupervised disentanglement methods, which justified the advantages of the StyleGAN architecture in disentanglement learning.
We then proposed Semi-StyleGAN that achieved near fully-supervised disentanglement with limited supervision (0.25\%$\sim$2.5\%) on complex high-resolution synthetic and real data. We also proposed new metrics to quantify the generator controllability. To the best of our knowledge, we were the first to reveal that there exists a trade-off between learning disentangled representations and controllable generation.
Besides, we extended Semi-StyleGAN to do semantic fine-grained image editing with better generalization to unseen images. 
Finally, we created two high-quality synthetic datasets to serve as new disentanglement benchmarks.

In the future, we want to apply Semi-StyleGAN to even larger-scale high-resolution real datasets. We are aware of the gender and racial biases in the CelebA dataset \cite{karkkainen2019fairface}, and thus hope to create better datasets and find other ways to address the algorithmic bias. Besides, It would be interesting to extend Semi-StyleGAN to the weakly-supervised, semi-supervised scenario, where the factors of variation are only partially observed. 

\vspace{-2mm}
\section*{Acknowledgement}
\vspace{-1mm}

Thanks to the anonymous reviewers for useful comments. We also thank Zhiding Yu, Anuj Pahuja, Yaosheng Fu, Tan Minh Nguyen and many others at Nvidia for helpful discussions on this work. WN and ABP were supported by IARPA via DoI/IBC contract D16PC00003.


\bibliography{example_paper}
\bibliographystyle{icml2020}


\newpage

\onecolumn

\appendix

\vspace{7mm}
\section*{\centering \Large Appendix}
\vspace{3mm}

\section{More details of Two New Datasets} \label{datasets}


\subsection{Examples of the Isaac3D Dataset}

\begin{figure} [ht]
	\centering
	\begin{subfigure}[b]{0.75\textwidth}
		\centering
		\includegraphics[width=\linewidth]{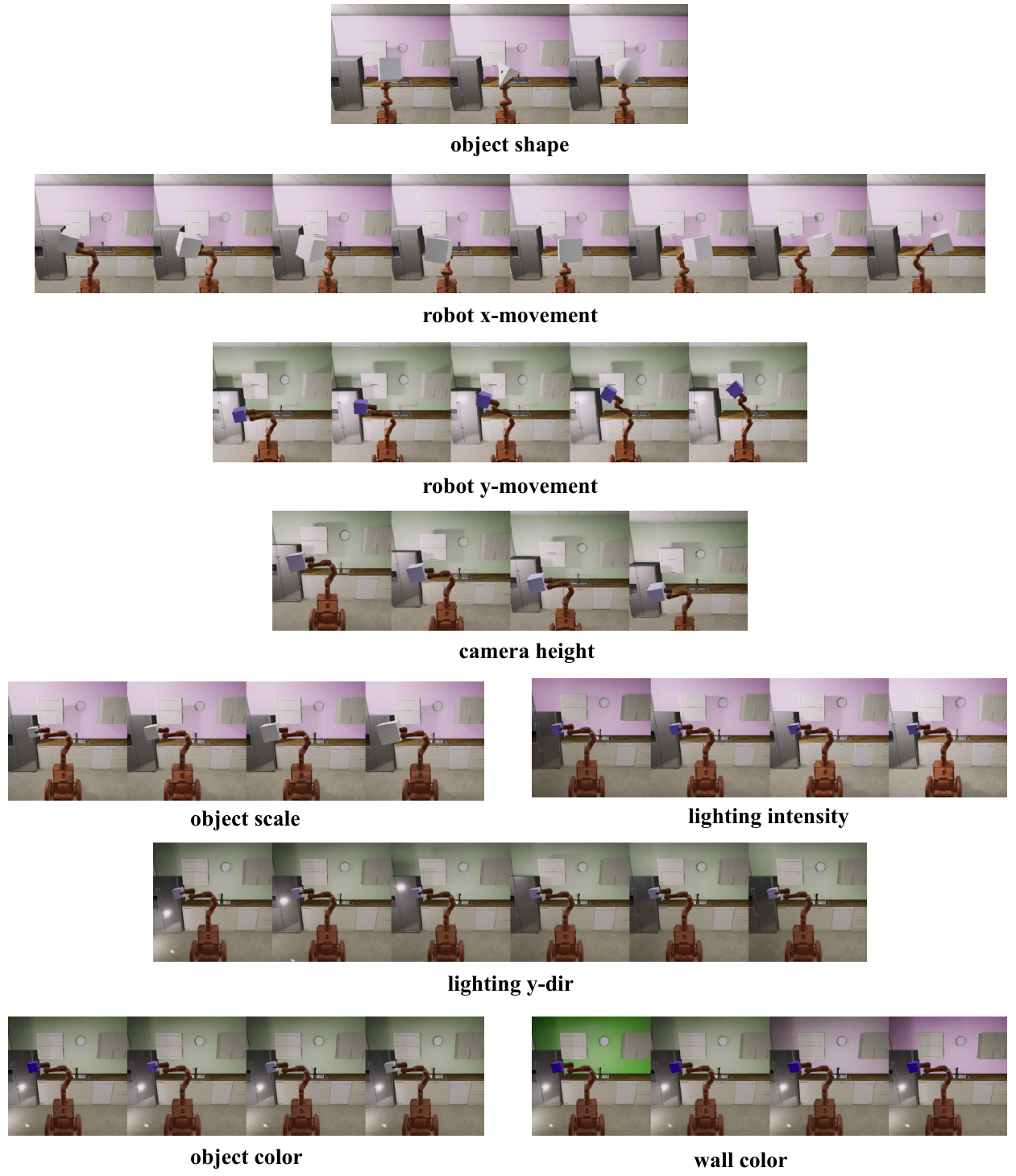}
	\end{subfigure}
	\caption{Examples of the Isaac3D dataset where we vary each factor of variation individually to see how each factor of variation changes with its corresponding ground-truth factor code.} 
	\label{isaac_samples_suppl}
\end{figure}

\newpage

\subsection{Examples of the Falcor3D Dataset}

\begin{figure} [h]
	\centering
	\begin{subfigure}[b]{0.60\textwidth}
		\centering
		\includegraphics[width=\linewidth]{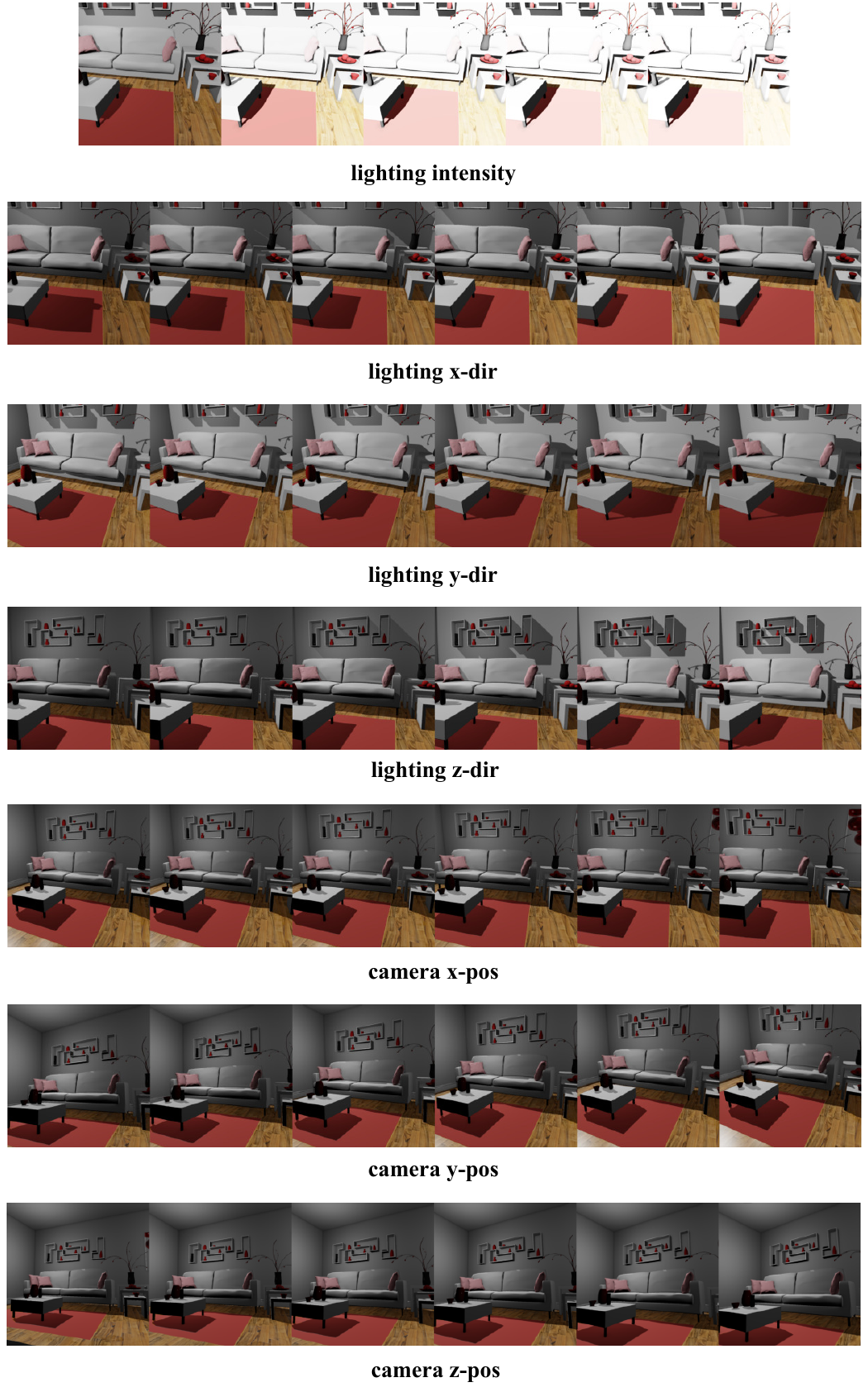}
	\end{subfigure}
	\caption{Examples in the Falcor3D dataset where we vary each factor of variation individually to see how each factor of variation changes with its corresponding ground-truth factor code.} 
	\label{falcor_samples_suppl}
\end{figure}

\newpage

\section{More results of Info-StyleGAN} \label{res_info-stylegan}

\subsection{Progressive Training for Disentanglement Learning} \label{progressive}

\begin{figure} [ht]
	\centering
	\begin{subfigure}[b]{0.38\textwidth}
		\centering
		\small
		\includegraphics[width=\linewidth]{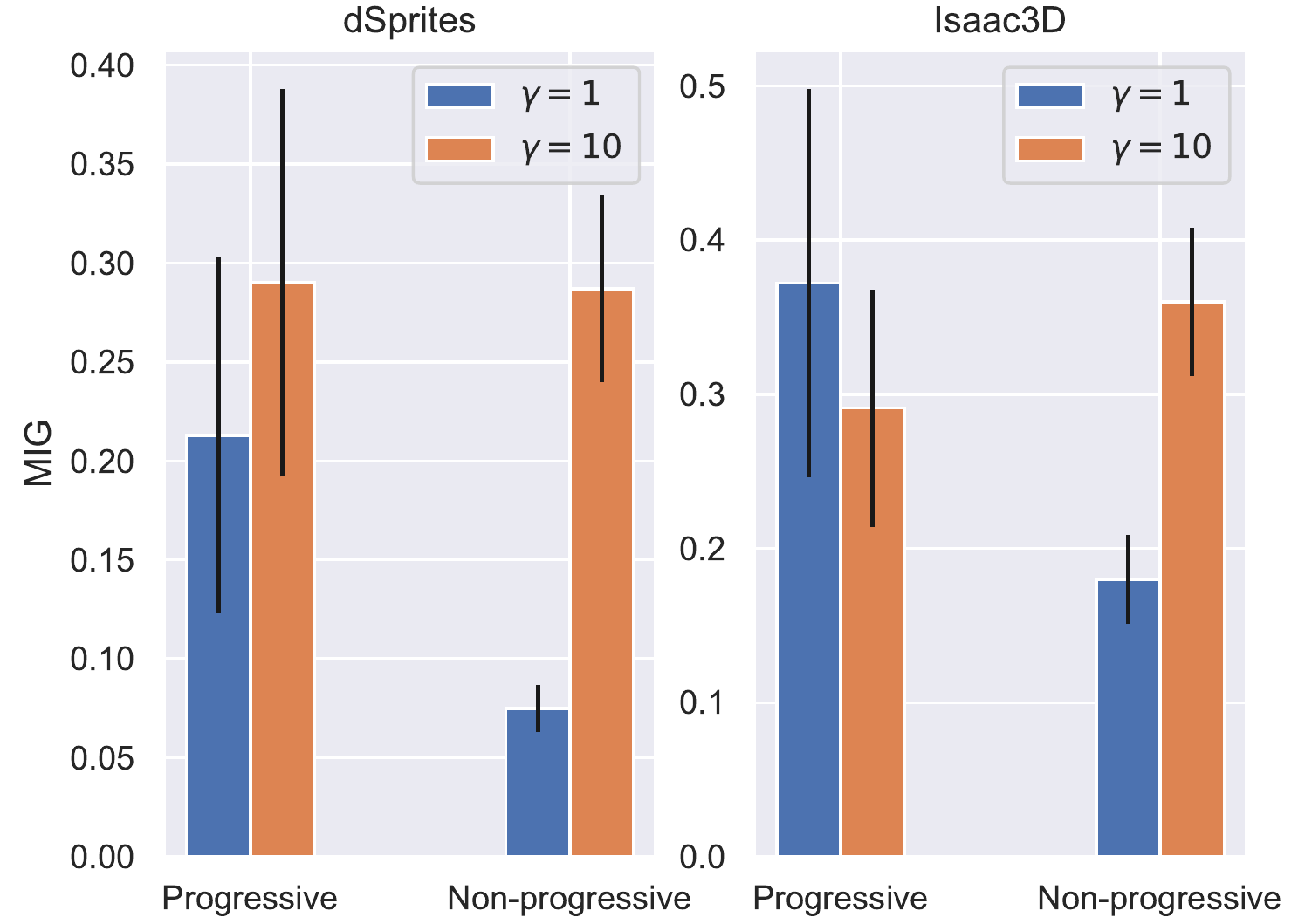}
	\end{subfigure}
	\caption{\small The impact of progressive training on the disentanglement learning with Info-StyleGAN. We can see that with progressive growing, the disentanglement quality tends to be better on both two datasets. Besides, the gap of average MIG scores with different values of the hyperparameter $\gamma$ is much smaller if the progressive growing is applied, which implies Info-StyleGAN with progressive growing seems to be less sensitive to  hyperparameters.}
	\label{his_pg}
\end{figure}

Progressive growing has been shown to improve the image quality of GANs \citep{karras2017progressive, karras2019style}, however, its impact on disentanglement learning remains unknown so far. Thus, we compare the MIG scores of Info-StyleGAN with progressive and non-progressive growing, respectively, on both dSprites and Isaac3D, and the results are shown in Figure \ref{his_pg}. We can see that with progressive growing, the disentanglement quality tends to be better on both two datasets. Besides, the gap of average MIG scores with different values of the hyperparameter $\gamma$ is much smaller if the progressive growing is applied, which implies Info-StyleGAN with progressive growing seems to be less sensitive to  hyperparameters. Therefore, unless stated otherwise, we use progressive growing for the GAN training in all the experiments.

\subsection{How to Get Info-StyleGAN* with Smaller Network Size} \label{samller_net} \label{match_net_size}

\begin{table}[pht!]
	\centering
	\begin{subtable}{0.45\linewidth}
		\centering
		\begin{tabular}{l}
			\toprule
			Mapping Network                                     \\ \hline 
			(FC $\times$ $n_{\text{mp}}$) $f_{\text{mp}}$ $\times$ $f_{\text{mp}}$
			\\ \hline
			\hline
			Synthesis Network
			\\ \hline
			(4$\times$4 Conv) 3$\times$3$\times$ $f_{\text{0}}$ $\times$ $f_{\text{0}}$ \\
			(4$\times$4 Conv) 3$\times$3$\times$ $f_{\text{0}}$ $\times$ $f_{\text{0}}$
			\\
			\hline
			(8$\times$8 Conv) 3$\times$3$\times$ $f_{\text{0}}$ $\times$ $f_{\text{0}}$ \\
			(8$\times$8 Conv) 3$\times$3$\times$ $f_{\text{0}}$ $\times$ $f_{\text{0}}$
			\\
			\hline
			(16$\times$16 Conv) 3$\times$3$\times$ $f_{\text{0}}$ $\times$ $f_{\text{0}}$ \\
			(16$\times$16 Conv) 3$\times$3$\times$ $f_{\text{0}}$ $\times$ $f_{\text{0}}$
			\\
			\hline
			(32$\times$32 Conv) 3$\times$3$\times$ $f_{\text{0}}$ $\times$ $f_{\text{0}}$ \\
			(32$\times$32 Conv) 3$\times$3$\times$ $f_{\text{0}}$ $\times$ $f_{\text{0}}$
			\\
			\hline
			(64$\times$64 Conv) 3$\times$3$\times$ $\frac{f_{\text{0}}}{2}$ $\times$ $\frac{f_{\text{0}}}{2}$ \\
			(64$\times$64 Conv) 3$\times$3$\times$ $\frac{f_{\text{0}}}{2}$ $\times$ $\frac{f_{\text{0}}}{2}$
			\\
			\bottomrule
		\end{tabular}
		\caption{Generator}
	\end{subtable}
	\quad
	\begin{subtable}{0.45\linewidth}
		\centering
		\begin{tabular}{l}
			\toprule
			(64$\times$64 Conv) 3$\times$3$\times$ $\frac{f_{\text{0}}}{2}$ $\times$ $\frac{f_{\text{0}}}{2}$ \\
			(64$\times$64 Conv) 3$\times$3$\times$ $\frac{f_{\text{0}}}{2}$ $\times$ $\frac{f_{\text{0}}}{2}$
			\\
			\hline
			(32$\times$32 Conv) 3$\times$3$\times$${f_{\text{0}}}$$\times$${f_{\text{0}}}$ \\
			(32$\times$32 Conv) 3$\times$3$\times$${f_{\text{0}}}$$\times$${f_{\text{0}}}$
			\\
			\hline
			(16$\times$16 Conv) 3$\times$3$\times$${f_{\text{0}}}$$\times$${f_{\text{0}}}$ \\
			(16$\times$16 Conv) 3$\times$3$\times$${f_{\text{0}}}$$\times$${f_{\text{0}}}$
			\\
			\hline
			(8$\times$8 Conv) 3$\times$3$\times$${f_{\text{0}}}$$\times$${f_{\text{0}}}$ \\
			(8$\times$8 Conv) 3$\times$3$\times$${f_{\text{0}}}$$\times$${f_{\text{0}}}$
			\\ \hline
			(4$\times$4 Conv) 3$\times$3$\times$${f_{\text{0}}}$$\times$${f_{\text{0}}}$ \\
			(4$\times$4 FC) (16$f_{\text{0}}$) $\times$ 64 \\
			(4$\times$4 FC) 64$\times$(1+ code\_length)
			\\
			\bottomrule
		\end{tabular}
		\caption{Discriminator / Encoder}
	\end{subtable}
	\caption{Generator and discriminator (or encoder) architectures in the implementation of Info-StylGAN for generating the image of resolution $128 \times 128$, where we use ``FC $\times n_{\text{mp}}$'' to denote that there are $n_{\text{mp}}$ dense layers in the given block, and use ``8$\times$8 Conv'' to denote the convolutional layer in the 8$\times$8 resolution block. Note that there is not the last block (i.e., 64$\times$64 Conv) in the generator and not the first block (i.e., 64$\times$64 Conv) in the encoder if we want to generate the image of resolution $64 \times 64$. For the original Info-StyleGAN, we have $n_{\text{mp}} = 8$, $f_{\text{0}} = 512$, $f_{\text{0}} = 512$. For Info-StyleGAN* on dSprites, we set $n_{\text{mp}} = 3$, $f_{\text{mp}} = 64$, $f_{\text{0}} = 64$ with 0.74M parameters in total. For Info-StyleGAN* on Downscaled Isaac3D, we set $n_{\text{mp}} = 3$, $f_{\text{mp}} = 256$, $f_{\text{0}} = 128$ with 3.44M parameters in total. }
	\label{tab:model_architecture}
\end{table}

\newpage

\subsection{Randomly Generated Samples of Baseline Models and Info-StyleGAN* (with Smaller Network Size)} \label{vae_samples}

\begin{figure} [ht]
	\centering
	\begin{subfigure}[b]{0.32\textwidth}
		\centering
		\includegraphics[width=\linewidth]{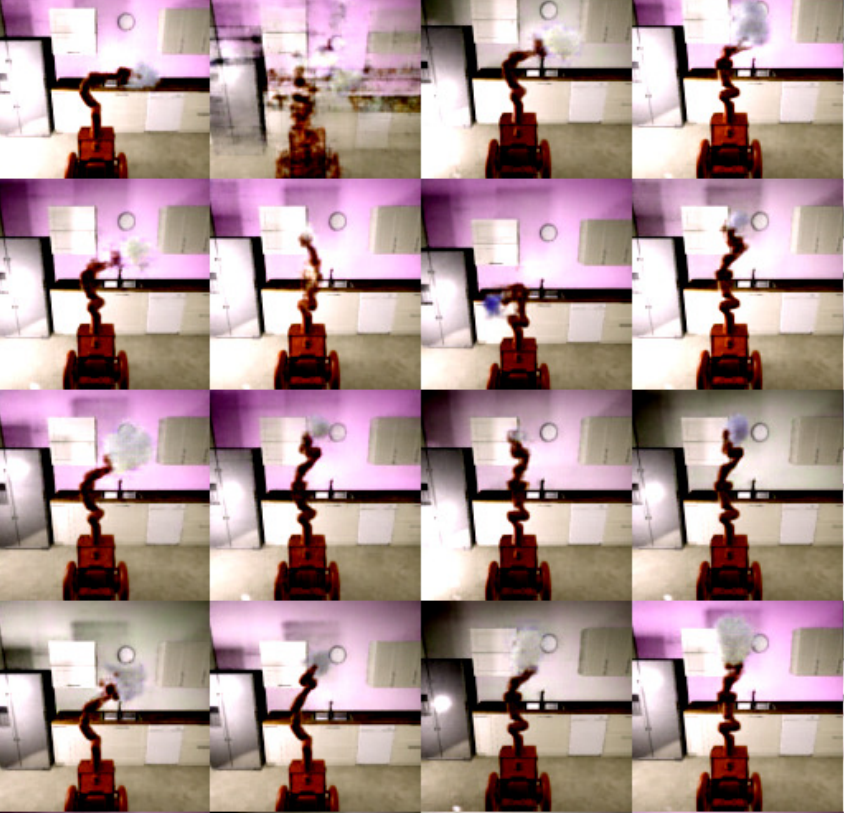}
		\caption{$\beta$-VAE (FID=120.3)}
	\end{subfigure}
	\begin{subfigure}[b]{0.32\textwidth}
		\centering
		\includegraphics[width=\linewidth]{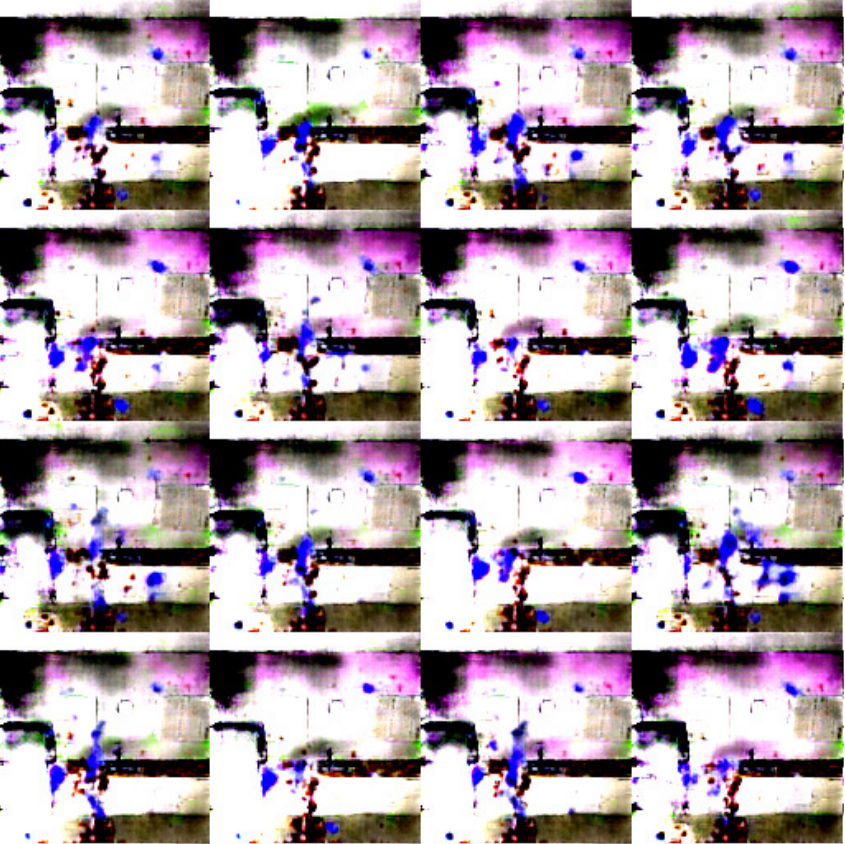}
		\caption{FactorVAE (FID=358.2)}
	\end{subfigure}
	\begin{subfigure}[b]{0.32\textwidth}
		\centering
		\includegraphics[width=\linewidth]{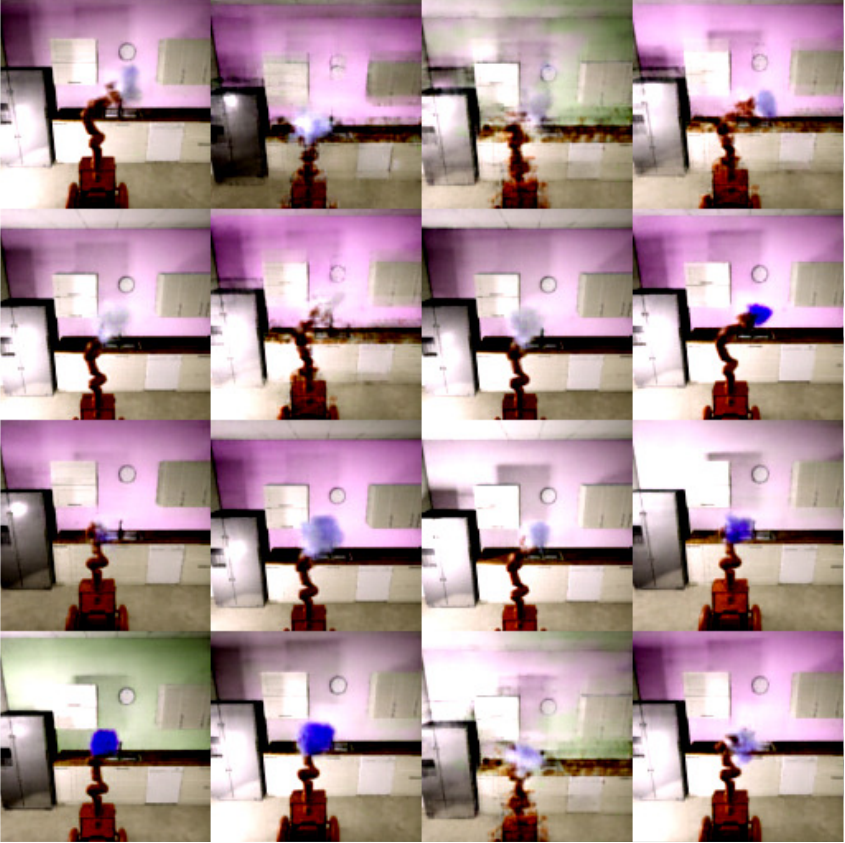}
		\caption{$\beta$-TCVAE (FID=143.8)}
	\end{subfigure}
	
	\begin{subfigure}[b]{0.32\textwidth}
		\centering
		\includegraphics[width=\linewidth]{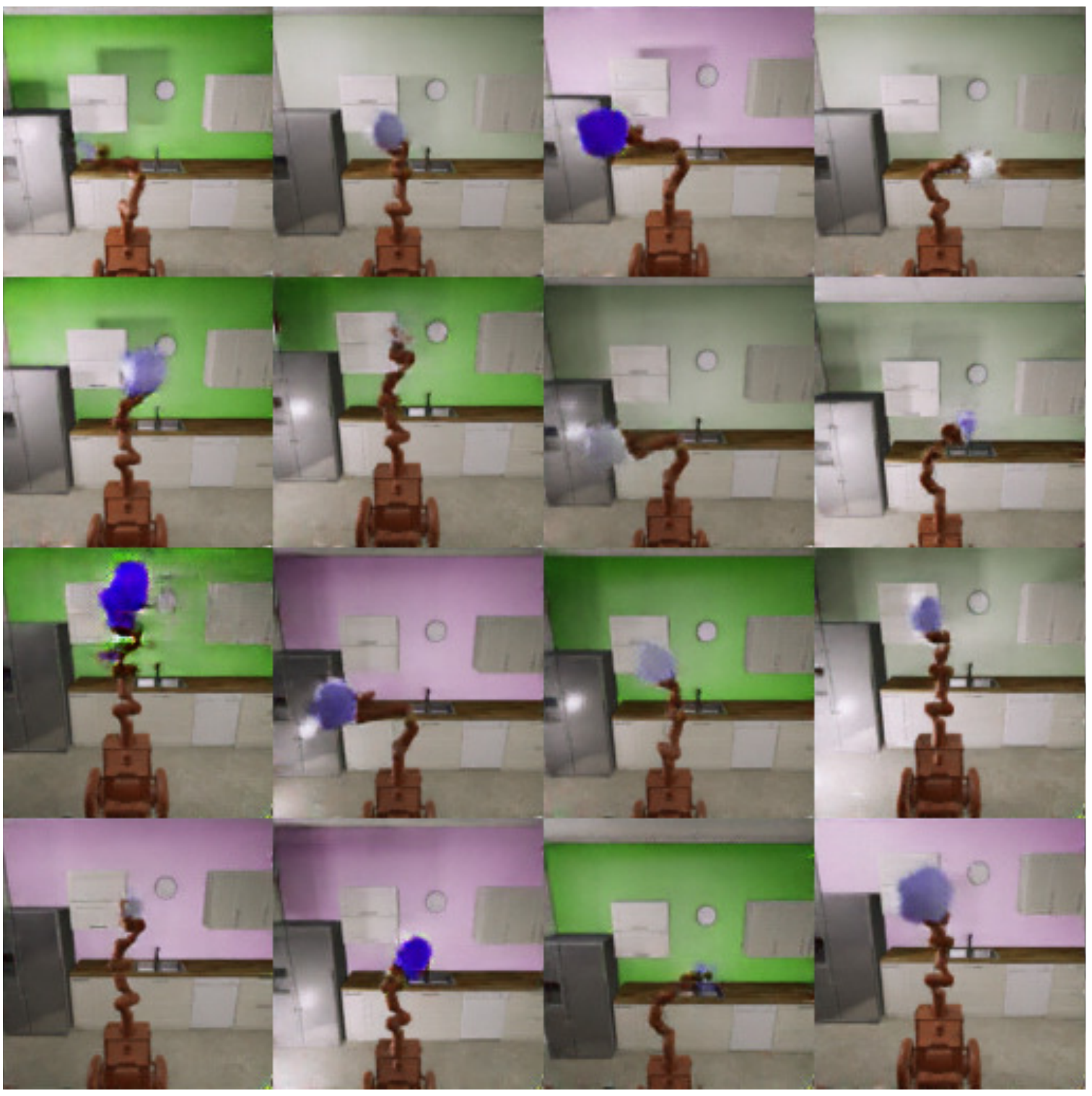}
		\caption{InfoGAN-CR (FID=73.43)}
	\end{subfigure}
	\begin{subfigure}[b]{0.32\textwidth}
		\centering
		\includegraphics[width=\linewidth]{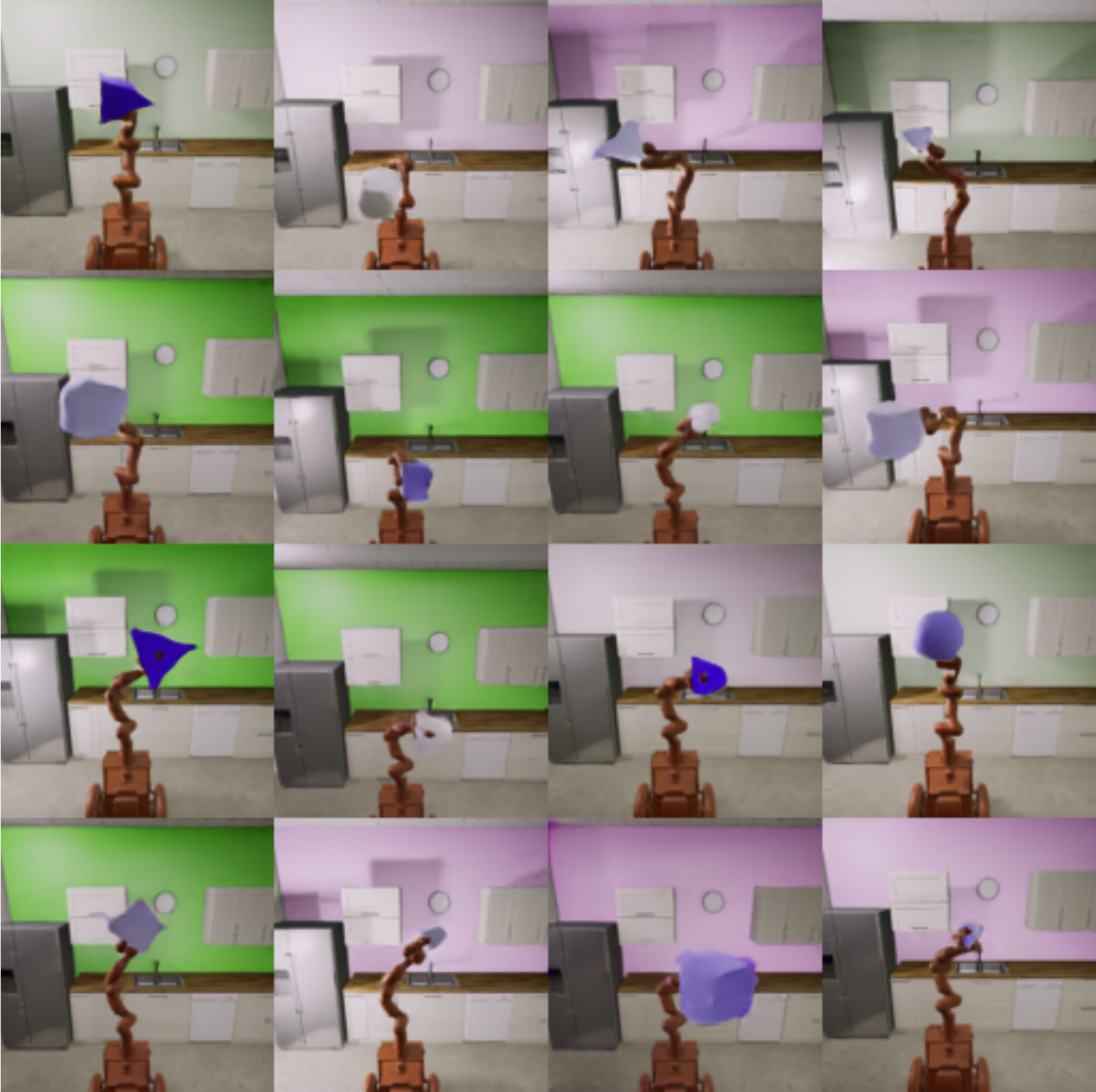}
		\caption{Info-StyleGAN* (FID=8.38)}
	\end{subfigure}
	\caption{\small {Randomly sampled images of baseline models and Info-StyleGAN* on (downscaled) Isaac3D of resolution 128x128. Note that for VAE-based models, we use the similar network architectures as in \cite{locatello2019challenging}, and Info-StyleGAN* denotes the smaller version of Info-StyleGAN, in which its number of parameters is similar to baseline models. We can see that compared with Info-StyleGAN* (of the same network size), the generated images of VAE-based models (i) tend to be quite blurry and of low quality, and (ii) fail to cover all the variations in the dataset. As a strong GAN baseline, InfoGAN-CR is also significantly worse than Info-StyleGAN in terms of image quality. The results demonstrate that our proposed dataset can serve as a new challenging benchmark for disentanglement learning, in particular regarding the much higher resolution, and larger variation of factors.} } 
	\label{latent_vae_suppl}
\end{figure}

\newpage

\subsection{Randomly Generated Samples of baseline models$^\dagger$ (with Larger Network Size) and Info-StyleGAN} \label{vae_samples_2}

\begin{figure} [ht]
	\centering
	\begin{subfigure}[b]{0.32\textwidth}
		\centering
		\includegraphics[width=\linewidth]{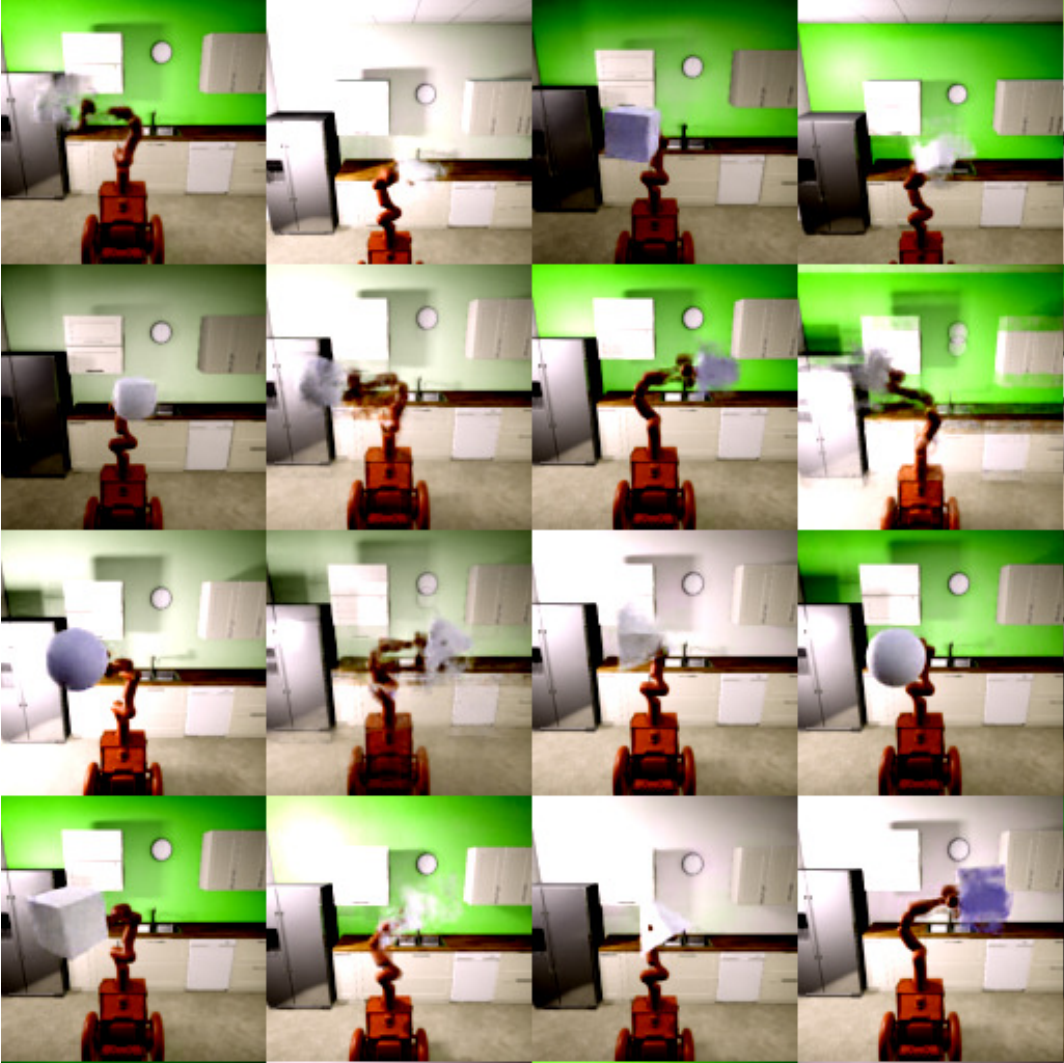}
		\caption{$\beta$-VAE$^{\dagger}$ (FID=60.71)}
	\end{subfigure}
	\begin{subfigure}[b]{0.32\textwidth}
		\centering
		\includegraphics[width=\linewidth]{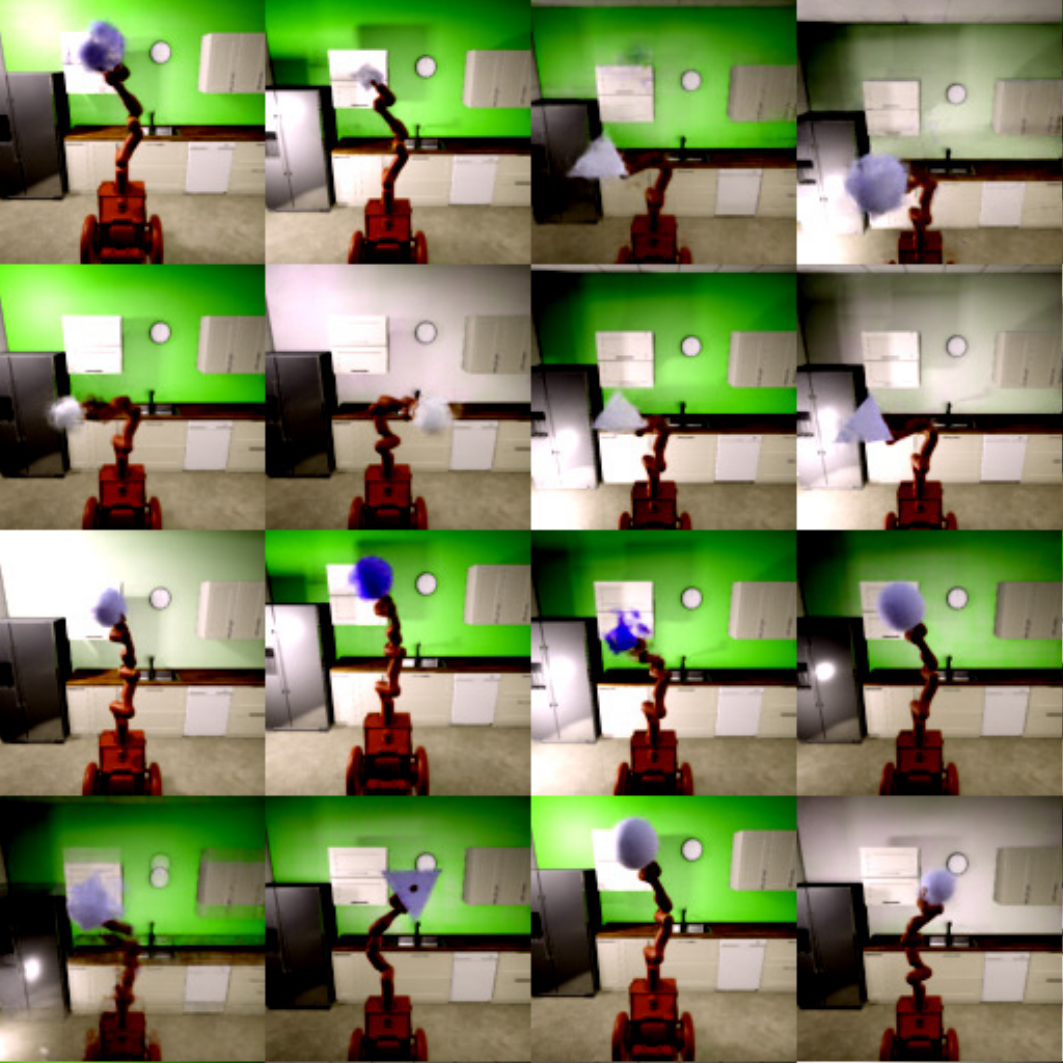}
		\caption{FactorVAE$^{\dagger}$ (FID=60.67)}
	\end{subfigure}
	\begin{subfigure}[b]{0.32\textwidth}
		\centering
		\includegraphics[width=\linewidth]{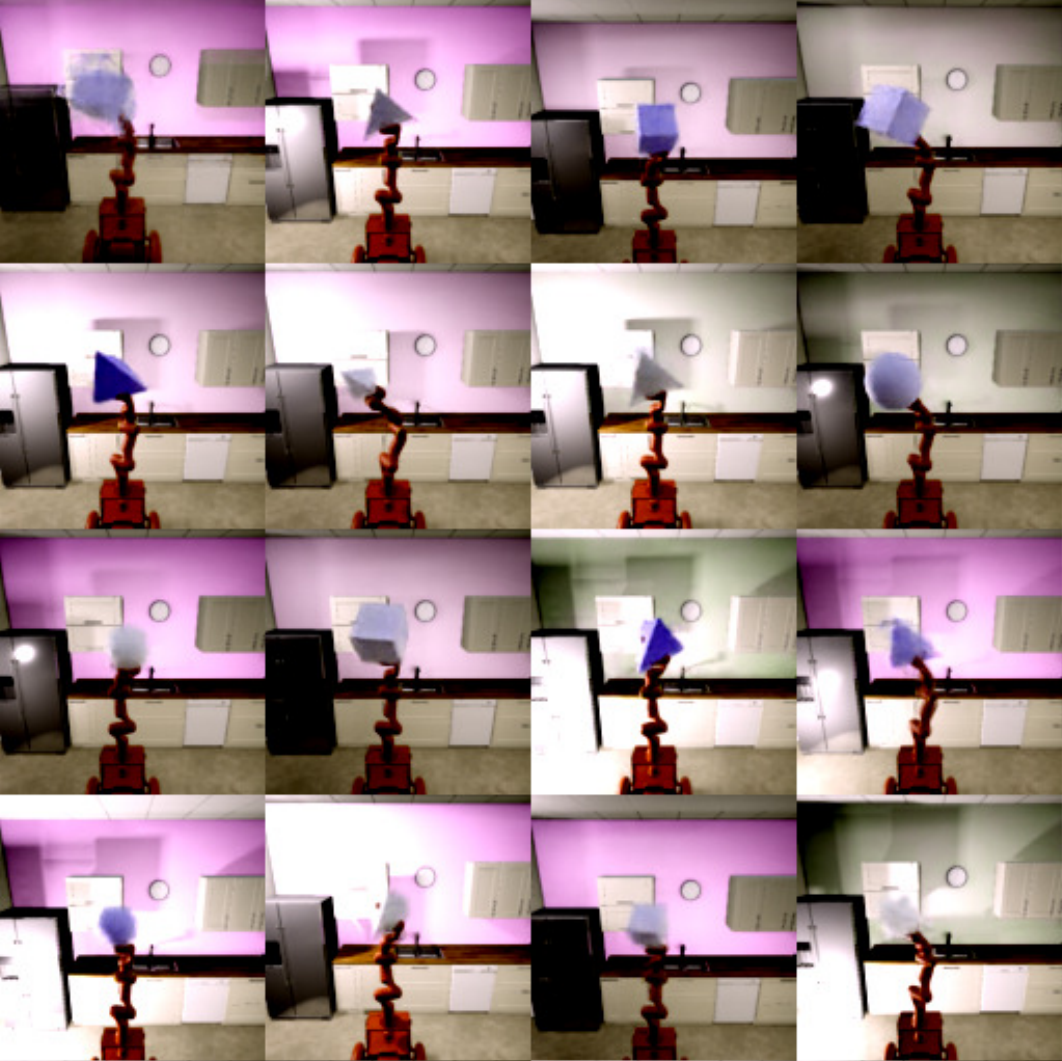}
		\caption{$\beta$-TCVAE$^{\dagger}$ (FID=77.48)}
	\end{subfigure}
	
	\begin{subfigure}[b]{0.32\textwidth}
		\centering
		\includegraphics[width=\linewidth]{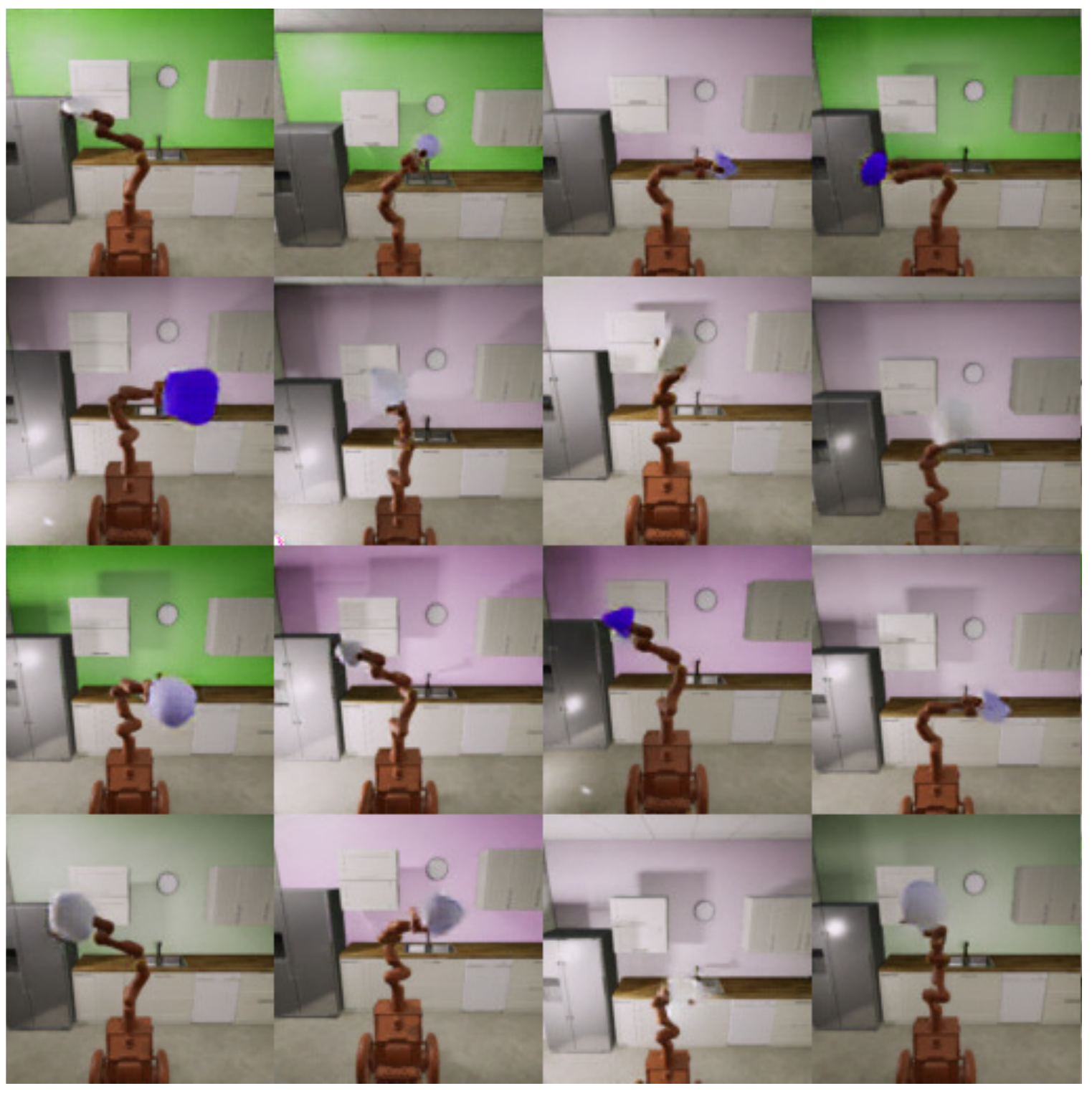}
		\caption{InfoGAN-CR (FID=30.41)}
	\end{subfigure}
	\begin{subfigure}[b]{0.32\textwidth}
		\centering
		\includegraphics[width=\linewidth]{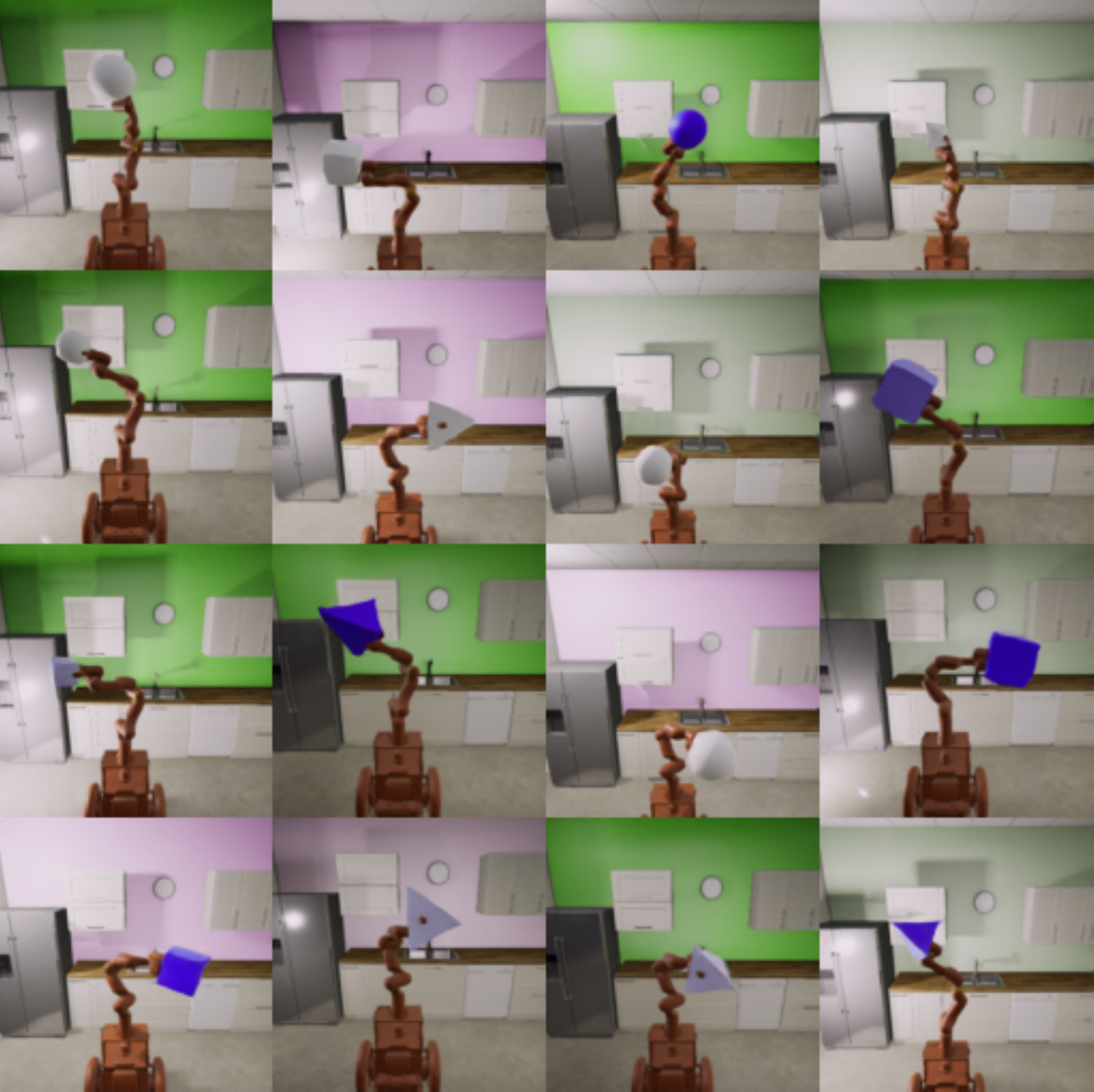}
		\caption{Info-StyleGAN (FID=2.50)}
	\end{subfigure}
	\caption{\small {Randomly sampled images of baseline models$^{\dagger}$ and Info-StyleGAN on (downscaled) Isaac3D of resolution 128x128. Note that for VAE-based models, we increase the number of featuremaps ($\times$8)  in each layer of the network architectures in \cite{locatello2019challenging}, so that their number of parameters is similar to Info-StyleGAN. We also apply the same operations for InfoGAN-CR to match the network size of In-StyleGAN. We can see that (i) the image quality gets better after increasing the network size, (ii) compared with Info-StyleGAN (of the same network size), the generated images of VAE-based models still have issues with blurriness and failure in capturing all variations, (iii) the generated images of InfoGAN-CR are better than VAEs but still worse than Info-StyleGAN.} } 
	\label{latent_vae_suppl_v1}
\end{figure}

\subsection{Other Experimental Settings} \label{exp_desc}

Our experiments are based on the StyleGAN implementation \citep{karras2019style}, where the GAN loss $L_{\text{GAN}}$, batch sizes, learning rates for both generator and discriminator, and other hyperparameters in the Adam optimizer and weights in each resolution black are all kept the same with \cite{karras2019style}, unless stated otherwise. 
Different from the original StyleGAN implementation, we do not apply truncation tricks. We also do not add noise inputs to introduce another randomness, as we consider the case where the factor code and latent $z$ will capture all the factors in the data. For all the quantitative results in the paper, we report the error bars by taking the mean and standard deviation of four runs with random seeds. For the implementation of evaluation metrics, we use 50K random sampled real images and fake images to calculate the FID score. We use 5K ground-truth observation-code pairs as training samples and 2K ground-truth observation-code pairs as test samples to evaluate the Factor score. We use 10K ground-truth observation-code pairs and 10K generated observation-code pairs to calculate the MIG and MIG-gen scores, respectively. We also use 1K ground-truth observation-code pairs and 1K generated observation-code pairs to calculate the L2 and L2-gen scores, respectively.

\newpage

\section{More Results of Semi-StyleGAN} \label{sec_semi}

\subsection{Semi-StyleGAN with 0.5\% of Labeled Data on Isaac3D with Resolution 512x512} \label{sec_semi_isaac3d}

\begin{figure} [ht]
	\centering
	\begin{subfigure}[b]{0.72\textwidth}
		\centering
		\includegraphics[width=\linewidth]{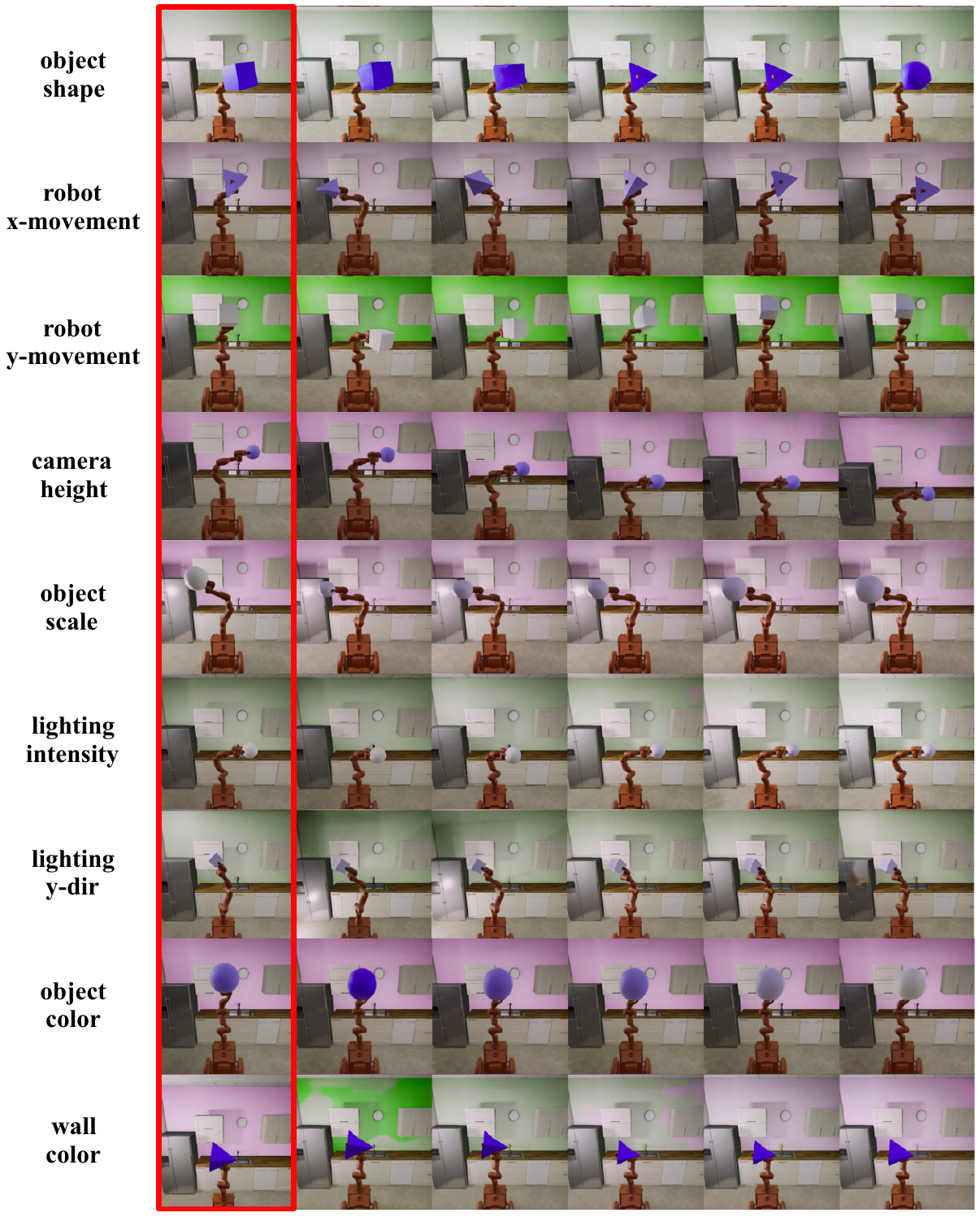}
	\end{subfigure}
	\caption{\small Latent traversal of Semi-StyleGAN on Isaac3D by using 0.5\% of the labeled data. Images in the first column (marked by red box) are randomly sampled real images of resolution 512x512 and the rest images in each row are their interpolations, respectively, by uniformly varying the given factor from 0 to 1.
		We can see that each factor changes smoothly during its interpolation without affecting other factors, and the interpolated images in each row visually look almost the same with their input image except the considered varying factor. Also, the image quality does not get worse during the interpolations.
	} 
	\label{latent_trav_isaac3d_all}
\end{figure}

\newpage

\subsection{Semi-StyleGAN with 1\% of Labeled Data on Falcor3D with Resolution 512x512} \label{sec_semi_falcor3d}

\begin{figure} [ht]
	\centering
	\begin{subfigure}[b]{0.82\textwidth}
		\centering
		\includegraphics[width=\linewidth]{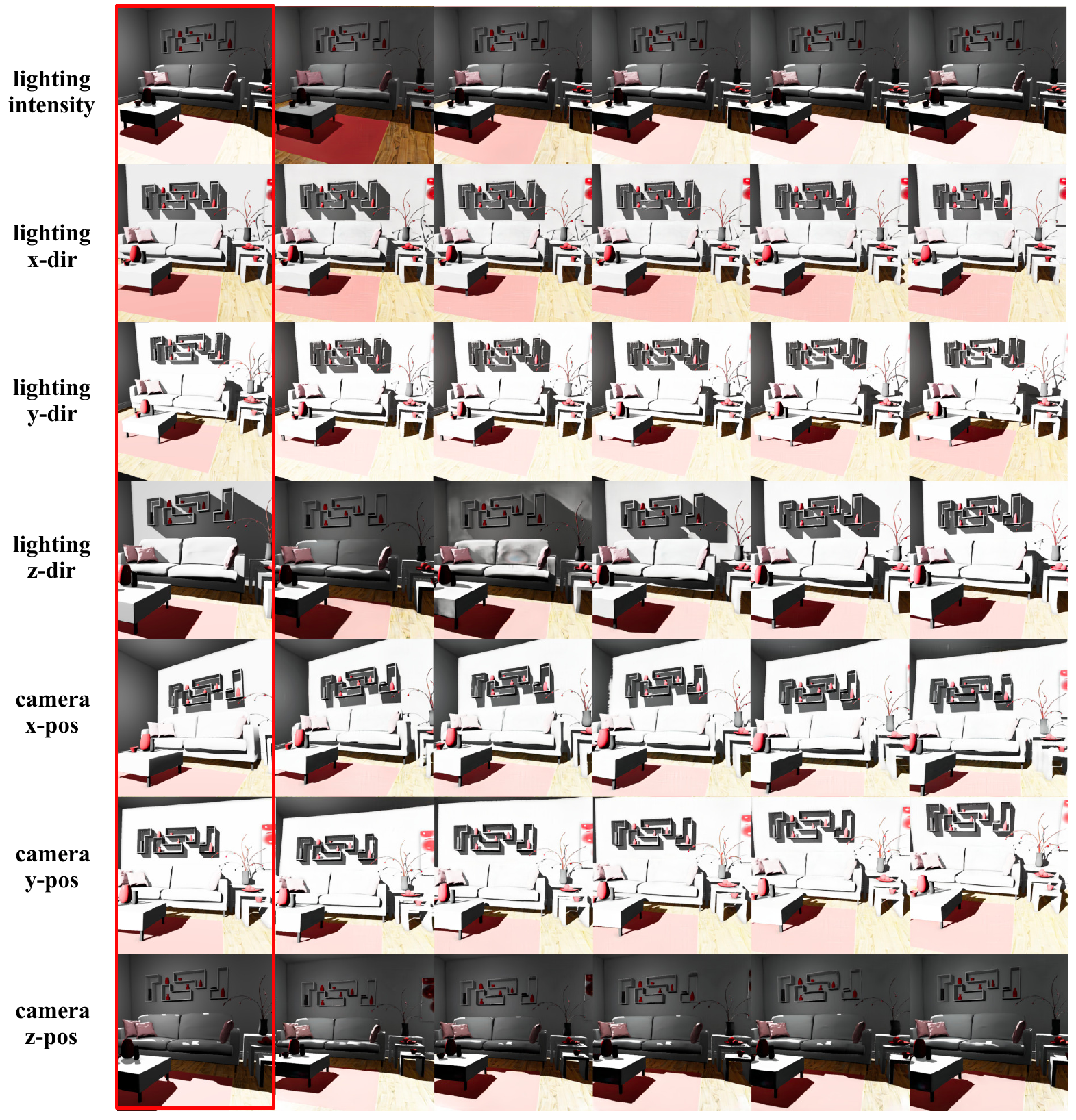}
	\end{subfigure}
	\caption{\small Latent traversal of Semi-StyleGAN on Falcor3D by using 0.5\% of the labeled data. Images in the first column (marked by red box) are randomly sampled real images of resolution 512x512 and the rest images in each row are their interpolations, respectively, by uniformly varying the given factor from 0 to 1.
		We can see that each factor changes smoothly during its interpolation without affecting other factors, and the interpolated images in each row visually look almost the same with their input image except the considered varying factor. Also, the image quality does not get worse during the interpolations.
	} 
	\label{latent_trav_falcor3d_all}
\end{figure}

\newpage 

\subsection{Semi-StyleGAN with 0.5\% of Labeled Data on CelebA with Resolution 256x256} \label{sec_semi_celeba}

\begin{figure} [ht]
	\centering
	\begin{subfigure}[b]{0.81\textwidth}
		\centering
		\includegraphics[width=\linewidth]{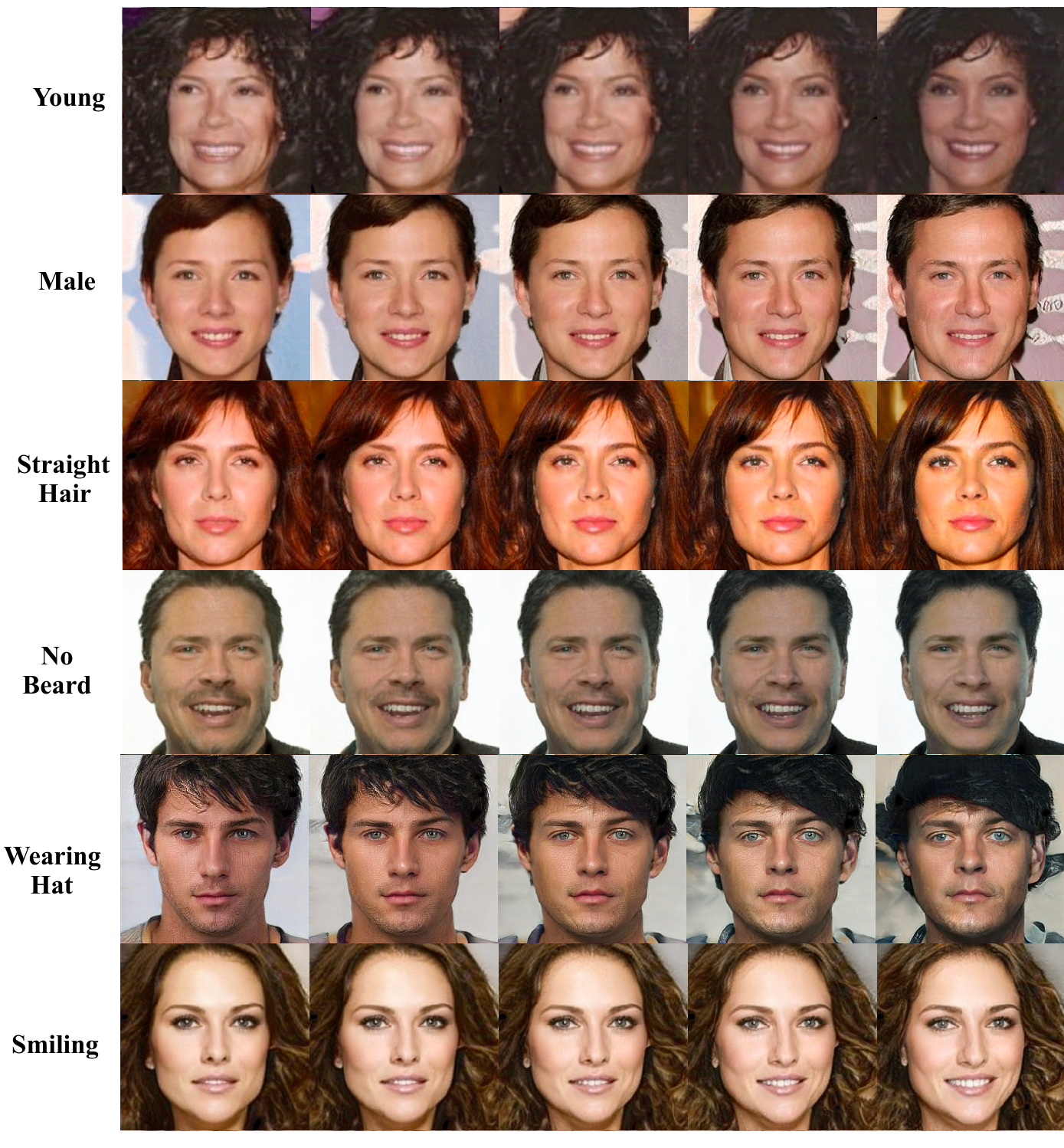}
	\end{subfigure}
	\caption{\small More latent traversal results of Semi-StyleGAN on CelebA with resolution 256x256 by using 0.5\% of the labeled data, where we control all 40 binary attributes at the same time. We can see that Semi-StyleGAN with only 0.5\% of the labeled data is capable of controlling the considered attributes. We note that the image background may also change slightly over interpolations of some attributions. We argue that it is because the other nuisance factors (i.e., those not in the set of considered 40 attributes) including background strongly confound the observed factors, which has been a common and difficult problem in high-dimensional partially observed latent variables models. We leave the investigation into how to further alleviate this confounding issue as the future work.
	} \label{semi_celeba_v2}
\end{figure}

\newpage

\begin{figure} [ht]
	\centering
	\begin{subfigure}[b]{0.81\textwidth}
		\centering
		\includegraphics[width=\linewidth]{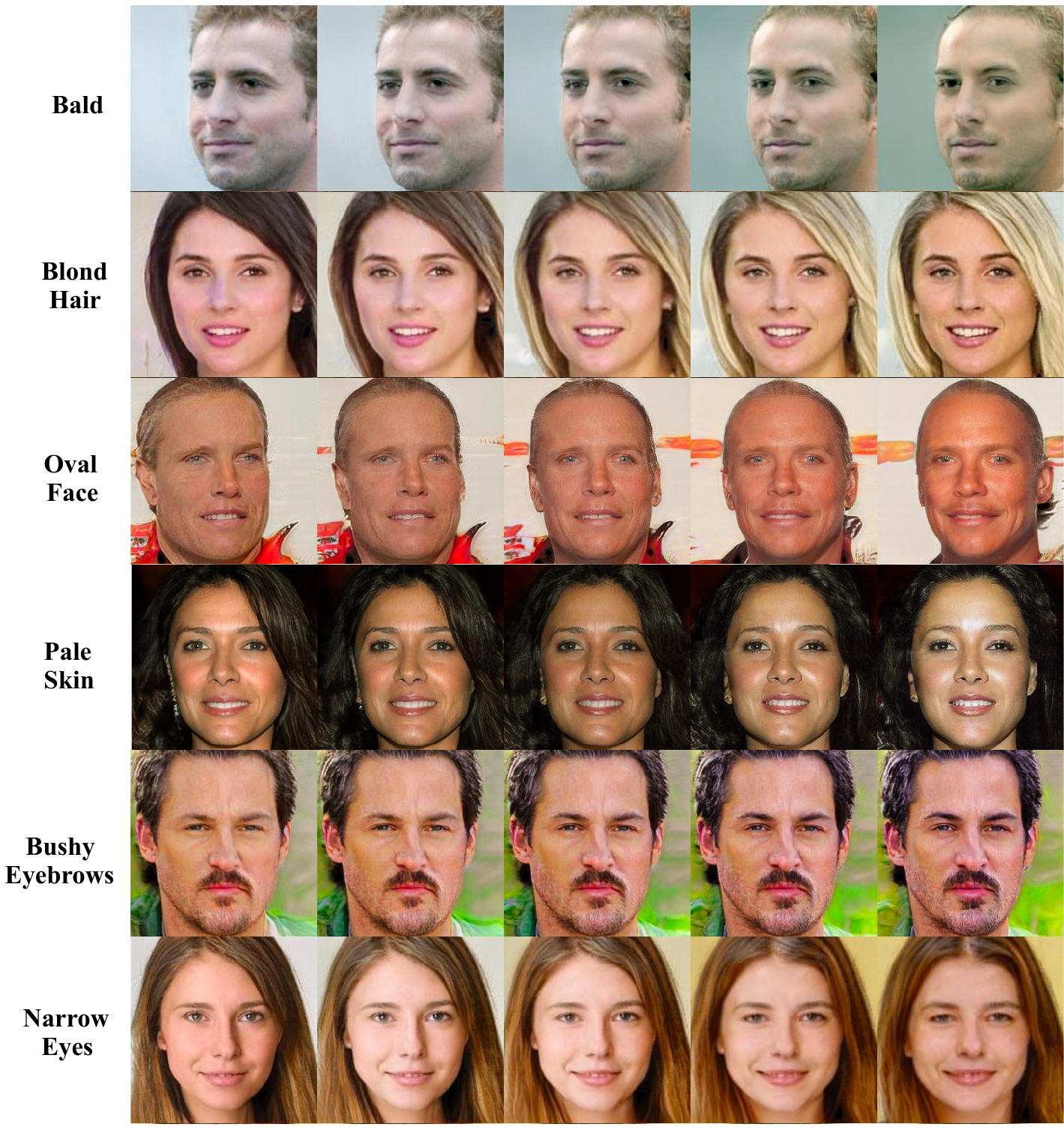}
	\end{subfigure}
	\caption{\small Latent traversal of Semi-StyleGAN on CelebA with resolution 256x256 by using 0.5\% of the labeled data, where we control all 40 binary attributes at the same time. We can see that Semi-StyleGAN with only 0.5\% of the labeled data is capable of controlling the considered attributes. We note that the image background may also change slightly over interpolations of some attributions. We argue that it is because the other nuisance factors (i.e., those not in the set of considered 40 attributes) including background strongly confound the observed factors, which has been a common and difficult problem in high-dimensional partially observed latent variables models. We leave the investigation into how to further alleviate this confounding issue as the future work.
	} \label{semi_celeba_v3}
\end{figure}

\newpage 

\section{More results on Semi-StyleGAN-\textit{fine}}
\label{sec_semi_ext}

\subsection{Semi-StyleGAN-\textit{fine} with 1\% of Labeled Data on Isaac3D Novel Images with Resolution 512x512} \label{sec_semi_ext_isaac3d}

\begin{figure} [ht]
	\centering
	\small
	\begin{subfigure}[b]{0.78\textwidth}
		\centering
		\includegraphics[width=\linewidth]{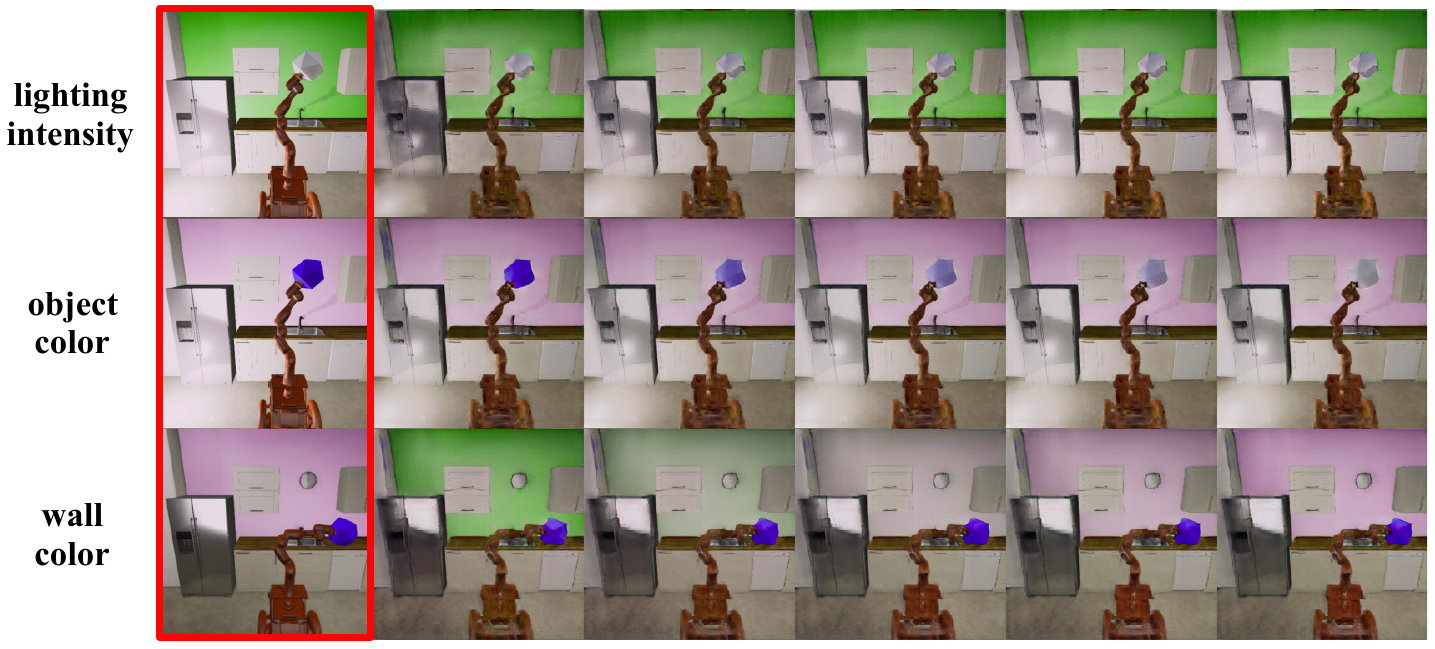}
	\end{subfigure}
	\caption{\small Generalization of Semi-StyleGAN-\textit{fine} with 1\% of the labeled data where we set $\phi=64$ and interpolate three fine styles: (lighting intensity, object color, wall color). In the test image, we shift the position of the robot arm to the right side, and also attach it with an unseen object (i.e., octahedron). 
		Images in the first column (marked by red box) are real novel images of resolution $512 \times 512$ and the rest images in each row are their interpolations, respectively, by uniformly varying the given factor from 0 to 1.
		We can see that Semi-StyleGAN-\textit{fine} with only 1\% of the labeled data is capable of controlling the considered fine-grained attributes without affecting the coarse-grained factors.  
	} 
	\label{ext_isaac3d_v2}
\end{figure}

\subsection{Semi-StyleGAN-\textit{fine} with 1\% of Labeled Data on CelebA Novel Images with Resolution 256x256} \label{sec_semi_ext_celeba}

\begin{figure} [ht]
	\setlength\belowcaptionskip{-10pt}
	\centering
	\small
	\begin{subfigure}[b]{0.89\textwidth}
		\centering
		\includegraphics[width=\linewidth]{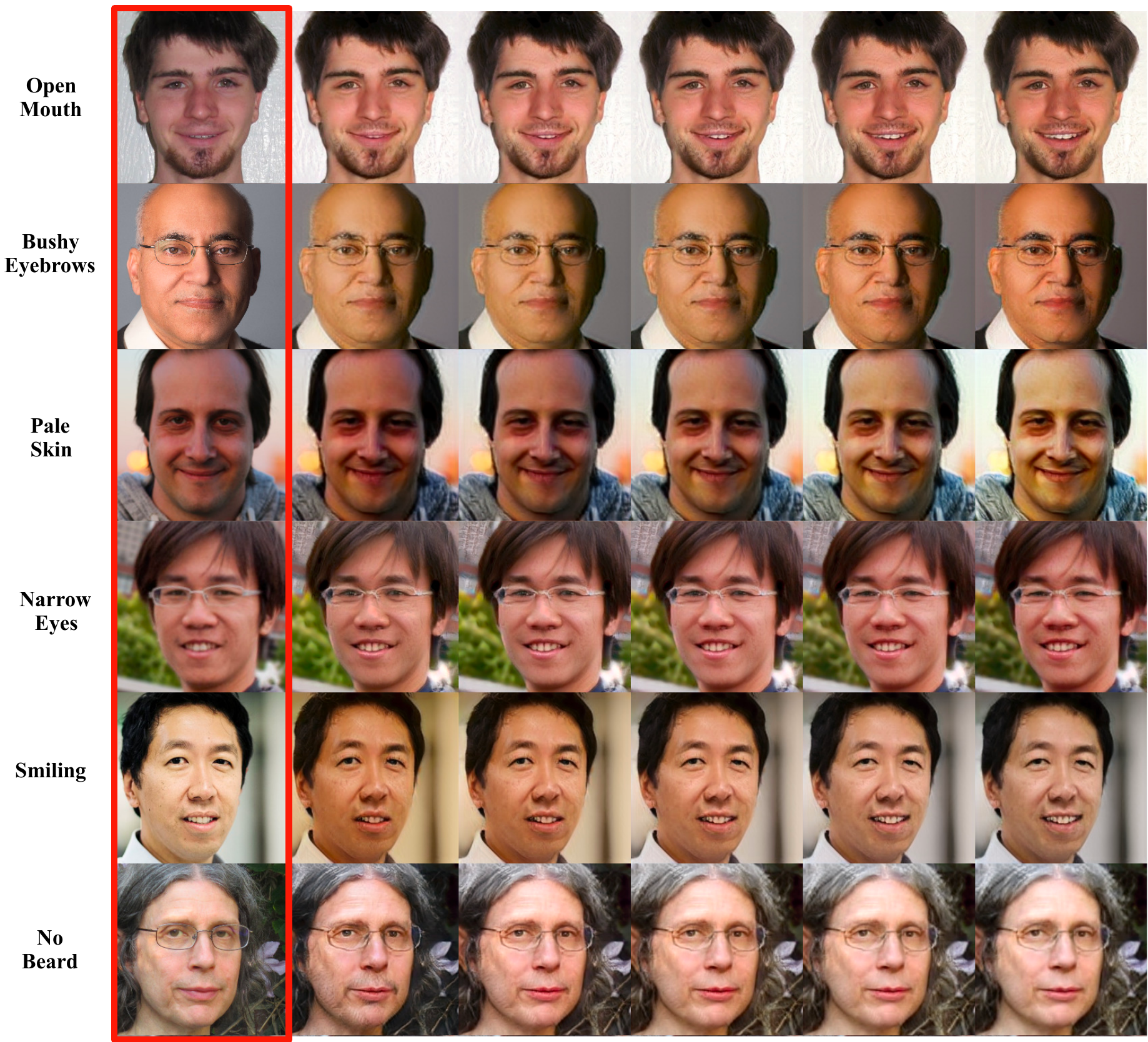}
	\end{subfigure}
	\caption{\small Generalized latent traversal results of Semi-StyleGAN-\textit{fine} trained on CelebA with 1\% of labeled data where we set $\phi=64$ and control the shown fine styles. 
		Images in the first column (marked by red box) are real novel images of resolution $256 \times 256$ and the rest images in each row are their interpolations, respectively, by uniformly varying the given factor from 0 to 1.
		We can see that Semi-StyleGAN-\textit{fine} with only 1\% of the labeled data is capable of controlling the considered fine-grained attributes without affecting the coarse-grained factors, in particular the personal identity.
	} 
	\label{ext_celeba_more}
\end{figure}


\end{document}